\makeatletter
\newenvironment{chapquote}[2][2em]
  {\setlength{\@tempdima}{#1}%
   \def\chapquote@author{#2}%
   \parshape 1 \@tempdima \dimexpr\linewidth-2\@tempdima\relax%
   \itshape}
  {\par\normalfont\hfill--\ \chapquote@author\hspace*{\@tempdima}\par\bigskip}
\makeatother

\pdfoutput=1

\documentclass[11pt,table,dvipsnames]{article}

\usepackage[final]{acl}
\usepackage{times}
\usepackage{latexsym}
\usepackage[utf8]{inputenc}

\usepackage[T1]{fontenc}

\usepackage[utf8]{inputenc}

\usepackage{microtype}

\usepackage{inconsolata}
\usepackage{microtype}
\usepackage{latexsym}
\usepackage{dsfont}
\usepackage{booktabs} 
\usepackage{CJKutf8}
\usepackage{graphicx}
\usepackage{bigstrut}
\usepackage{multirow}
\usepackage{amsmath}
\usepackage{multirow, makecell, caption}
\usepackage{floatrow}
\newfloatcommand{capbtabbox}{table}[][\FBwidth]
\usepackage{xcolor}
\usepackage[normalem]{ulem}
\usepackage{amssymb}
\usepackage{amsfonts}
\usepackage{multicol}
\usepackage{tipa}
\usepackage{amsmath}
\usepackage{bm}
\usepackage[ruled,linesnumbered]{algorithm2e}
\usepackage{tikz}
\usepackage{paralist}
\usepackage{IEEEtrantools}
\usepackage[switch]{lineno}
\usepackage{enumitem}
\usepackage{multirow, makecell, caption}
\usepackage{arydshln}
\usepackage{verbatim}
\usepackage[normalem]{ulem}
\usepackage{amssymb}
\usepackage{amsfonts}
\usepackage{multicol}
\usepackage{amssymb}
\usepackage{pifont}
\usepackage{csquotes}
\usepackage{xspace}
\usepackage{paralist}
\usepackage{mdwlist}
\usepackage{subcaption}
\usepackage{makecell}
\usepackage{array}
\usepackage{bm}
\usepackage{colortbl}
\usepackage{dsfont}
\usepackage{url}
\usepackage{booktabs} 
\usepackage{graphicx}
\usepackage{tabularx}
\usepackage{amsmath}
\usepackage{bbm}
\usepackage{pgfplots}
\usepackage{soul} 
\usepackage{listings}
\usepackage[tableposition=top]{caption}
\usepackage{color,xcolor} 

\definecolor{Ground}{RGB}{255,184,55}
\definecolor{Rice}{RGB}{251,248,238}
\definecolor{Dirt}{RGB}{191,169,115}
\definecolor{Pink}{RGB}{226,184,176}
\definecolor{Violet}{RGB}{163,148,170}
\definecolor{mygray}{RGB}{226, 226, 226}

\newcolumntype{g}{>{\columncolor{Ground!10}}c}
\newcolumntype{d}{>{\columncolor{Dirt!10}}c}
\newcolumntype{f}{>{\columncolor{Pink!10}}c}
\newcolumntype{v}{>{\columncolor{Violet!10}}c}

\makeatletter
\def\adl@drawiv#1#2#3{%
        \hskip.5\tabcolsep
        \xleaders#3{#2.5\@tempdimb #1{1}#2.5\@tempdimb}%
                #2\z@ plus1fil minus1fil\relax
        \hskip.5\tabcolsep}
\newcommand{\cdashlinelr}[1]{%
  \noalign{\vskip\aboverulesep
           \global\let\@dashdrawstore\adl@draw
           \global\let\adl@draw\adl@drawiv}
  \cdashline{#1}
  \noalign{\global\let\adl@draw\@dashdrawstore
           \vskip\belowrulesep}}
\makeatother

\newcommand{\xmark}{\ding{55}}
\newcommand{\cmark}{\ding{51}}
\newcommand{\dataset}{\textsc{LifeChoice}\xspace}

\newcommand{\method}{\textsc{CharMap}\xspace}

\newcommand{\customcite}[2]{\hyperlink{cite.#2}{#1}}

\title{Character is Destiny: Can Role-Playing Language Agents Make Persona-Driven Decisions?}

\author {
    Rui Xu$^{1}$, 
    Xintao Wang$^{1}$,
    Jiangjie Chen$^{1}$,
    Siyu Yuan$^{1}$,
    Xinfeng Yuan$^{1}$,\\
    \textbf{
    Jiaqing Liang$^{1}$,
    Zulong Chen$^{2}$,
    Xiaoqing Dong$^{2}$,
    Yanghua Xiao$^{1}$} \\
    $^1$Fudan University \quad
    $^2$Alibaba Group \quad \\
    \texttt{\{ruixu21, xtwang21, syyuan21, xfyuan23\}@m.fudan.edu.cn} \\
    \texttt{\{jjchen19, liangjiaqing, shawyh\}@fudan.edu.cn} \\
    \texttt{\{zulong.czl, xiaoqing.dongxq\}@alibaba-inc.com}
}

\begin{document}
\begin{CJK}{UTF8}{gbsn}

\maketitle

\begin{abstract}
Can Large Language Models~(LLMs) simulate humans in making important decisions? 
Recent research has unveiled the potential of using LLMs to develop role-playing language agents (RPLAs),  mimicking mainly the knowledge and tones of various characters. 
However, imitative decision-making necessitates a more nuanced understanding of personas. 
In this paper, we benchmark the ability of LLMs in persona-driven decision-making. 
Specifically, we investigate whether LLMs can predict characters' decisions provided by the preceding stories in high-quality novels.  
Leveraging character analyses written by literary experts, we construct a dataset \dataset comprising 1,462 characters' decision points from 388 books.
Then, we conduct comprehensive experiments on \dataset, with various LLMs and RPLA methodologies.
The results demonstrate that state-of-the-art LLMs exhibit promising capabilities in this task, yet substantial room for improvement remains. 
Hence, we further propose the \method method, which adopts persona-based memory retrieval and significantly advances RPLAs on this task, achieving 5.03\% increase in accuracy. 
Resources are available at \url{https://github.com/airaer1998/LifeChoice}.


\begin{figure}[t]
    \centering
    \includegraphics[width=0.95\linewidth]{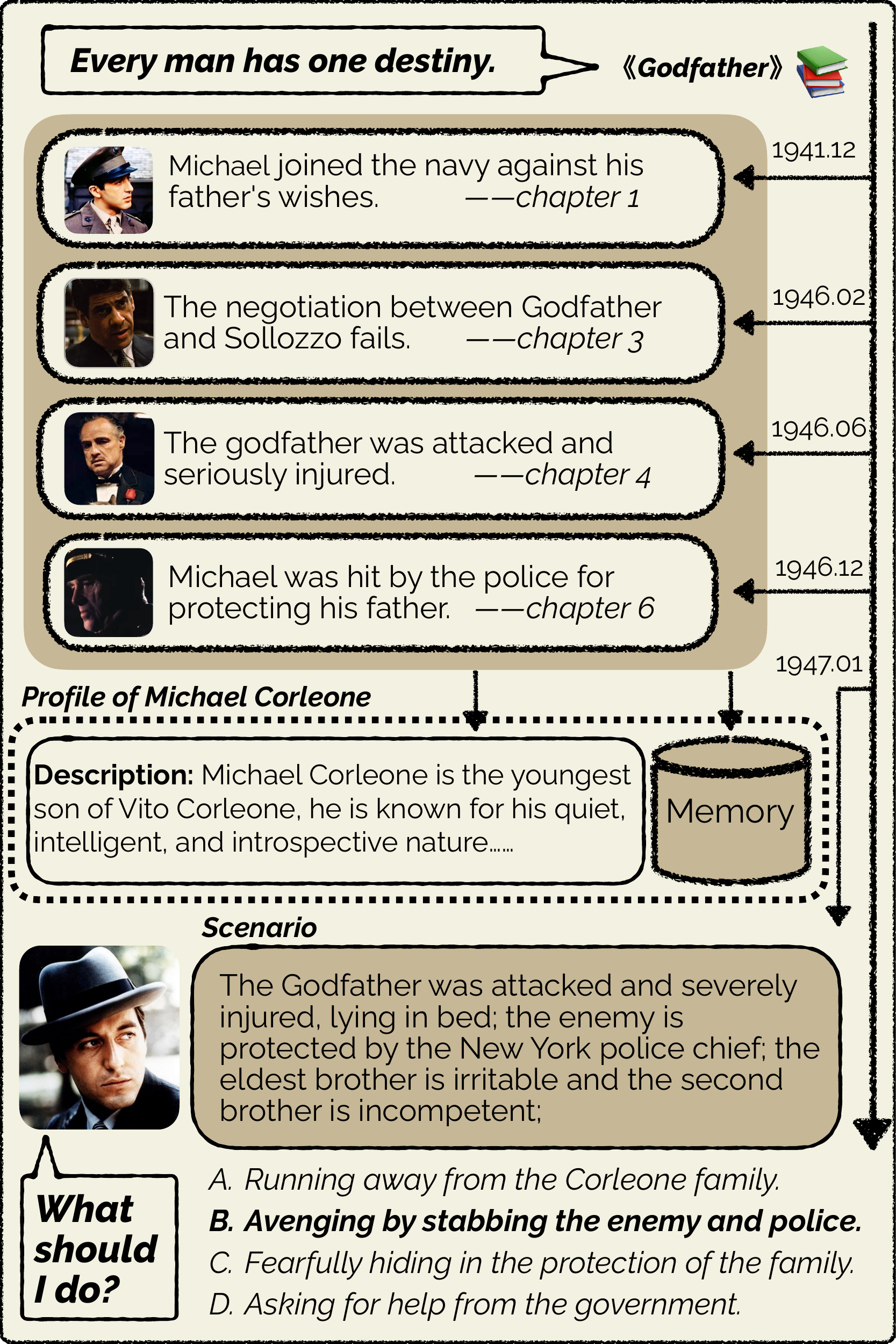}
    \caption{
    An example of \dataset. Given a character, a decision point and the preceding context, RPLAs are expected to reproduce the original decision. 
    Typically, RPLAs are constructed by parsing the context into the character's description and memory. 
    }
    \label{fig:main}
\end{figure}




\end{abstract}

\section{Introduction}
\label{sec:intro}
\begin{chapquote}{\normalsize{\customcite{\textit{Julius Caesar}}{ShakespeareJulius}. Act 1, Scene 2.}}
The fault, dear Brutus, is not in our stars, but in ourselves, that we are underlings.
\end{chapquote}


With the recent advancements in large language models~(LLMs)~\cite{openai2023gpt4,touvron2023llama}, Role-Playing Language Agents~(RPLAs) have emerged as a flourishing field of AI applications and research~\cite{chen2024persona}. 
RPLAs are LLM-based AI systems that simulate assigned personas, reproducing their tones, knowledge, personalities and even decisions~\cite{park2023generative,gao2024retrievalaugmented, wang2023does, xie2024can}. 
They emulate various characters across extensive applications, including fictional characters in chatbots and video games~\cite{wang2023rolellm,wang2023does}, as well as digital clones~\citep{gao2023livechat} or personalized assistants~\cite{Xu_2022,salemi2024lamp} for real-world individuals. 


Can RPLAs reliably make decisions that align with their personas, as humans do? 
This question is vital for the practical usage of RPLAs, yet remains underexplored. 
Previous studies primarily investigate RPLAs' character fidelity in terms of their tones~\citep{wang2023rolellm} and  knowledge~\citep{shao2023character}, which could be readily replicated by existing RPLAs via style imitation and knowledge retrieval. 
However, these features are relatively superficial compared with the underlying thinking and mindset of characters. 
Recent efforts~\citep{wang2023does} study the personality fidelity of RPLAs, but 
they fail to capture the nuances and dynamics of characters' mindsets. 
Hence, it remains an understudied question whether RPLAs could simulate persona-driven decisions, which challenges their comprehensive understanding of the personas and reasoning about unobserved behaviors. 

In this paper, we systematically study the capability of RPLAs to simulate persona-driven decisions, based on characters from high-quality novels.
In high-quality novels, characters' life choices are carefully plotted and aligned with their personas. 
Hence, we introduce the \dataset dataset, which 
evaluate whether RPLAs can faithfully reproduce the characters' life choices in the narratives.  
Specifically, \dataset comprises 1,462 character decisions from 388 novels, leveraging expert-written character analyses.
Each sample is presented as a multiple-choice question with the preceding context before the decision point. 
As depicted in Figure \ref{fig:main}, RPLAs are expected to identify and reason over relevant knowledge about the characters to simulate their decisions.  
The construction of \dataset primarily involves three steps: decision point selection, multiple-choice question construction, and manual examination.

Compared with previous methods for RPLA evaluation, our task and dataset benefit from higher-quality data and are more challenging. 
First, our questions and decisions are well-designed and closely aligned with the personas, since they are sourced from well-crafted  narratives.
Hence, our data establish solid ground truth for simulating characters' persona-driven decisions.
Second, our task is more challenging as it requires RPLAs to comprehensively understand and reason based on the personas, including their knowledge, experiences, and personalities. 
Specifically, \dataset poses the following challenges:
1) \textit{Long-context understanding}, where RPLAs need to identify sparse relevant motivations from massive character contexts. 
2) \textit{Temporal intelligence}, where RPLAs should intelligently adapt to the dynamic evolution of characters and environments.  
3) \textit{Intricate motives}, where RPLAs are required to reason through complex and entangled backgrounds and motives to arrive at the decisions.

We conduct extensive experiments to evaluate RPLAs on \dataset.
Our experiments cover various LLMs and different RPLA frameworks, including memory-enhanced agents, long-context LLMs, and our proposed method \method towards better simulation of persona-driven decisions. 
The results demonstrate that 
existing RPLAs have shown a promising accuracy of up to 62.92\% on \dataset. Furthermore, \method significantly enhances RPLAs on this task, achieving an accuracy of 67.95\%, which exceeds previous baselines by 5.03\%. 
However, compared to the human performance of 92.01\%, there is still significant room for improvement. 
Meanwhile, we observe that both well-summarized character descriptions and accurate memory retrieval are crucial for RPLAs. 

In summary, our contributions include:
\begin{itemize}[noitemsep]
    \item We propose to explore RPLAs' ability in 
     simulating persona-driven decisions, which is crucial for future RPLA applications and challenges existing RPLAs. 
    \item We delicately craft \dataset, the first benchmark for persona-driven decisions of RPLAs, based on characters' life choices from high-quality novels. 
    Besides, we propose \method, which adopts persona-based memory retrieval for better decision-making of RPLAs. 
    \item Based on \dataset, we conduct extensive experiments. The results demonstrate the promising performance of RPLAs in decision simulation.  
    Then, we analyze and compare methodologies for RPLA development, and show the effectiveness of \method. 
\end{itemize}

\section{Related Work}
\label{sec:related}
\paragraph{Character Role-Playing}
Early research on character-related studies focuses on character understanding. 
\citet{brahman2021let} attempts to predict a specific character through the text of the novel. 
\citet{yu2022few} provides dialogues from movie scripts for the model to examine and then asks it to identify the character who speaks each passage. 
With the enhancement of model abilities, some work attempts to make the model simulate complex role-playing.
\citet{li2023chatharuhi} analyzes 32 anime characters using 54k dialogues and personality traits. 
They use sentence embeddings for dialogue selection and evaluation. 
\citet{zhou2023characterglm} uses identity, interests, and relationships, collecting AI behaviors for imitation and using character data for fine-tuning. They evaluate model consistency and linguistic style. 
\citet{wang2023rolellm} creates a dataset for script characters and evaluates role-playing quality based on speaking style imitation and role-specific knowledge.
These studies make a chatbot for a certain character, but they focus more on imitating the character from the perspective of dialogue, which is a shallow imitation. 
Our study aims to evaluate RPLAs from the perspective of behavior and decision-making, which further challenges the model's understanding of the roles.

\paragraph{Personal LLM assistants}
With the rapid development of artificial intelligence technology, there are now many personal intelligent agents embedded in mobile devices, providing personalized services through analyzing user data and equipment~\cite{kaplan2019siri,hoy2018alexa}.
These agents can model the user's profile and preferences through the user's historical data~\cite{gurrin2014lifelogging,dodge2007outlines}, such as extracting personality from the user's record text~\cite{majumder2017deep,vstajner2020survey}, reading emotions from the user's image data~\cite{jaiswal2020facial,zad2021emotion}, modeling preferences from historical interaction information~\cite{tang2019akupm,li2018automated}, and pushing notifications from smart phones~\cite{li2018automated}. These memories can enhance the model's decision-making and reasoning, bringing a better personal experience for users. However, obtaining real user memory data is difficult and violates privacy. We model characters from historical data in high-quality novel texts, allowing the model to restore the real choices in the storyline based on the previous text, providing the first benchmark for the wide testing of personal intelligent agents.

\section{Dataset and Task Setups}
\label{sec:dataset}
\subsection{Dataset Construction} 
\label{sec:dataset_construction}


\begin{table}[t]
  \centering
  \small
    \begin{tabularx}{\linewidth}{X}
    \toprule
    \textbf{Book}: Les Misérables \\
    \textbf{Character}: Jean Valjean \\
    \textbf{Context}: \\
    In 1815 Monsieur Charles-François-Bienvenu Myriel was Bishop of Digne. He was then......Jean Valjean reflections gave him a sort of frightening aspect. He was subject to one of those violent inner tearings, which was not unknown to him. \\
    \textbf{Scenario}: \\
    In the courtroom, an innocent man was wrongfully accused, because he bore a resemblance to Jean Valjean. If Jean Valjean did not come forward, this innocent man would be sent to the gallows in his place. At this time, Jean Valjean had transformed his identity and become a respected town mayor, and he had also adopted a young girl named Cosette, with whom he had a new life. \\
    \textbf{Question}: \\
    \textcolor{teal} {You will play the role of Jean Valjean. What will you choose to do when you discover that man is about to be convicted due to being mistaken for you?} \\
    \textbf{Options}: \\
    A. Keep silent, letting an innocent person take the punishment in one's place. \\
    B. Persuade the person to run away, in order to protect both from the disaster of jail. \\
    \textcolor{brown} {C. Go to court and reveal the truth, sacrificing oneself to save the innocent person.} \\
    D. Look for legal loopholes, trying to save both the person and oneself. \\
    \midrule
    \textbf{Correct Answer}: C \\
    \textbf{Motivation}: \\
    \textit{[Values and Beliefs]} Jean Valjean is a person who values honesty and justice, possessing a strong sense of morality and righteousness. He decides to turn himself in to save another innocent person, fulfilling his inner need for morality and justice. \\
    \bottomrule
    \end{tabularx}
  \caption{Case study of \dataset. A complete set of data includes book, character, scenario, question, options, correct answer, motivation, and input.}
  \label{tab:case}
\end{table}

\begin{table*}[t]
\centering
\small
\begin{tabular}{lcccc}
\toprule
\textbf{Dataset} & \textbf{Source} & \textbf{Context Length} & \textbf{Task Format} & \textbf{Has Explanation} \\
\midrule
\texttt{TVSHOWGUESS} & TV show transcripts & ～50k & Character Identification & \xmark \\
\texttt{ROCStories} & Commonsense short stories
 & ～100 & Character Behavior Prediction & \xmark  \\
\texttt{LiSCU} & Literature
 & ～1000 & Character Identification & \xmark \\
\dataset & Literature
 & ～150k & Character Behavior Prediction & \cmark
 \\ \bottomrule
\end{tabular}
\caption{Comparison between \dataset and previous
character understanding benchmarks:
data source, context length, task format, and whether the benchmark has explanations.}
\label{tab:dataset}
\end{table*}

We construct a comprehensive dataset called \dataset. 
As shown in Table~\ref{tab:case}, the sample for each decision point includes the preceding context $p$ from the original book, the current scenario $s$, a question $q$ outlining a decision faced by that character $c$, a list of options $a = \{a_i\}_{i=1}^4$, the correct answer $y$, and the motivation $m$ explaining the character’s choice. 
Our data is sourced from the website \textit{Supersummary}\footnote{\url{https://www.supersummary.com/}}, which provides three pieces of content written by literary experts: \textit{key character descriptions}, \textit{chapter summaries}, and \textit{book analyses}. 
We contact the website and obtain authorization to use the data for academic research. 
The dataset construction comprises the following three main steps:


\paragraph{Selecting Decision Points}
To prevent data leakage, we first filter novels on the site using the following criteria:
(1) The narrative must exclude non-fiction genres like biographies or documentary literature.
(2) The narrative perspective must be in the first or third person.
(3) The progression of narrative time should be linear, avoiding stories with complex timelines or flashbacks.
(4) Exclude overly popular books, as measured by a high number of reviews on literary review websites. 
For each book that passes these filters, we provide GPT-4 with content written by literary experts, requesting it to output each key character's life choice decision points and their corresponding gold motivations. 
Additionally, we also ask GPT-4 to output the chapter numbers related to the decision based on the extracted motivations.
As shown in the example in Figure~\ref{fig:main}, the literary expert's analysis of the book suggests that \textit{Michael Corleone}'s motivation for choosing to assassinate the enemy includes both avenging his father and witnessing the collusion between the police and the enemy, which exposes him to the darker side of the government. We then identify two corresponding chapters in the original book based on these motivations, providing more refined data for constructing multiple-choice questions.

\paragraph{Constructing Multiple-Choice Questions}
We input the content written by literary experts and the corresponding chapters identified based on motivation into GPT-4. 
Our goal is to generate multiple-choice questions that capture the complexity of the characters' decision-making processes. 
The correct option reflects the decision made by the characters in the original books, whereas the distractors are designed to be plausible for  an arbitrary person.
As shown in the example in Figure~\ref{fig:main}, \textit{Michael Corleone} can ask for help from the government because he was once a Navy officer who trusted the government. 
However, in the preceding text, \textit{Michael} witnesses the dark side of the government, so he ultimately chooses to stab the police.

\paragraph{Manual Examination}

We invite ten native English-speaking university students to filter the data and pay them according to local minimum wage standards.
We supply the annotators with content written by literary experts and the multiple-choice questions, asking them to assess whether the model-created questions are challenging and reasonable.
They are also tasked with filtering out data they deem low quality. 

Ultimately, we collect 1,462 characters from 388 books and their corresponding life choices. Table \ref{tab:case} shows a complete data example. The specific prompts and more detailed data construction process can be found in Appendix \ref{sec:appendix_prompt}. 

\begin{figure}[t]
    \centering
    \includegraphics[width=\linewidth]{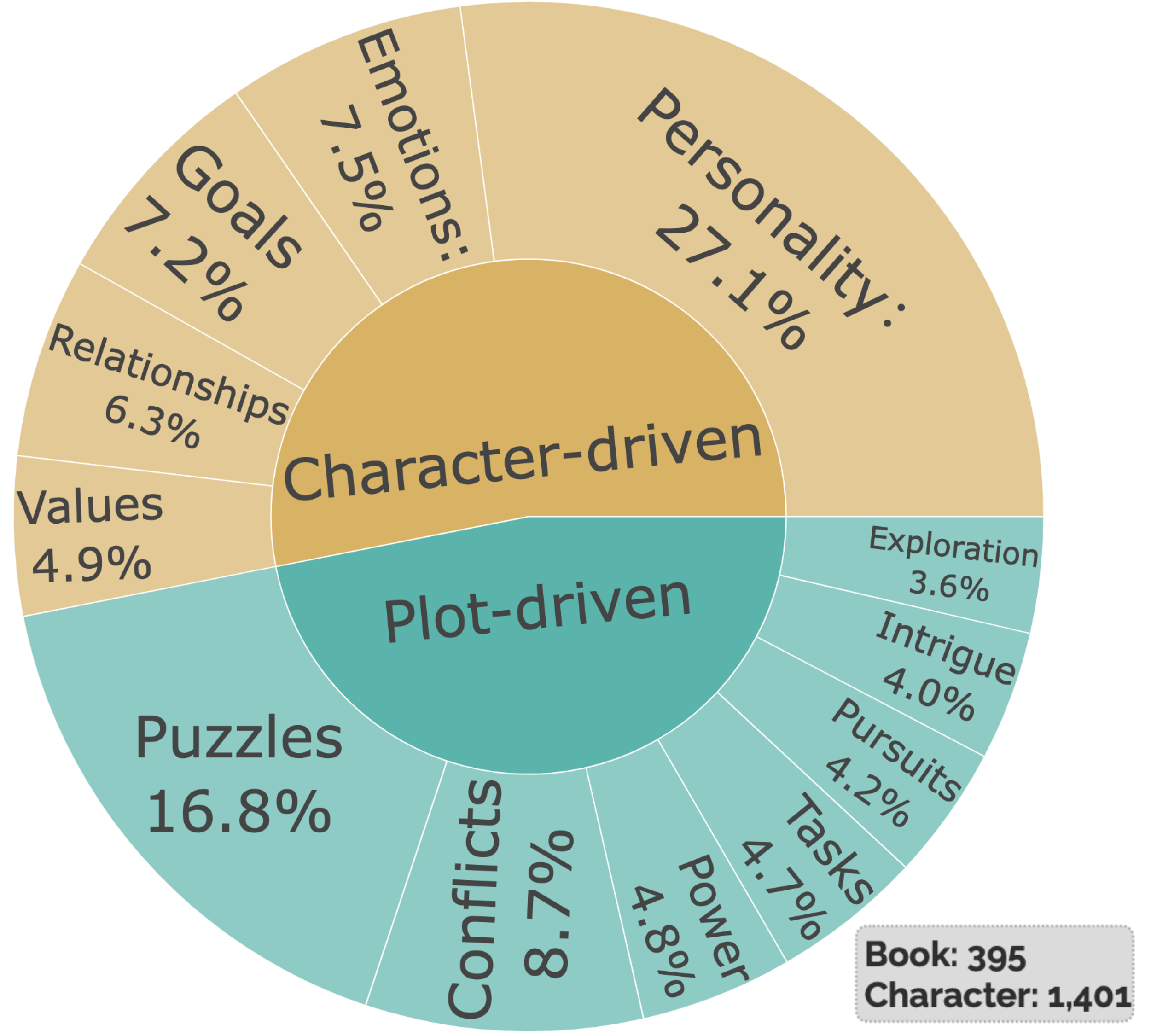}
    \caption{Statistics of motivation types in \dataset, with the first words for each motivation type.}
    \label{fig:data}
\end{figure}

\subsection{Dataset Analysis}
\label{sec:dataset_analysis}
We refer to the drama theory of Aristophanes~\cite{sommerstein2013aristophanes,silk2002aristophanes} as the system prompt and use GPT-4 to classify the motivations for character decisions into two meta-motivations and several accompanying sub-motivations:

\paragraph{Character-driven Motivation}
Character-driven behavior revolves around the character's inner world, personality, and transformation. Sub-motivations of character-driven behavior include \textit{Personality and Traits}, \textit{Emotions and Psychological State}, \textit{Social Relationships}, \textit{Values and Beliefs}, and  \textit{Desires and Goals}.







\paragraph{Plot-driven Motivation}
Plot-driven behavior stems from a series of external events and conflicts unfolding. Characters often react passively within a larger narrative structure, with their actions led by external events. Sub-motivations of plot-driven behavior include \textit{External Conflicts}, \textit{Tasks and Goals}, \textit{Puzzles and Secrets}, \textit{Pursuits and Escapes}, \textit{Exploration and Discovery}, \textit{Power and Control}, and \textit{Intrigue and Betrayal.}









Note that each topic is assigned one category of motivation. Figure \ref{fig:data} shows the proportion of different motivations. Detailed introductions for each sub-motivation are in Appendix \ref{appendix:category_2}.

\subsection{Task Setups} 
\label{sec:task_setups}
This task can be formulated as $P(y|x)$. Given the input $x=(p, s, c, q, a)$, the RPLA needs to identify the correct choice $y$ that aligns with the character's decision in the narrative.
For evaluation, we directly use the accuracy of multiple-choice question answering. 
As shown in Table \ref{tab:dataset}, compared to other character understanding tasks, \dataset requires understanding the character through a more extended context to make decisions.
RPLAs must locate relevant information related to the current scene in vast personal data.
This behavior demands a more profound understanding of the characters.

\section{Experiments}
\label{sec:experiments}

\begin{table}[t]
\centering  
\small
\begin{tabular}{llcc}
\toprule
\multirow{2}{*}{\makecell{\textbf{Profile}\\\textbf{Construction}}} & 
\multirow{2}{*}{\makecell{\textbf{Role-Playing}\\\textbf{Model}}} & 
\multirow{2}{*}{\makecell{\textbf{ACC}}} & \multirow{2}{*}{\makecell{\textit{\textbf{+motivation}}}} \\
& & & \\
\midrule
\rowcolor[gray]{0.95} \multicolumn{4}{c}{\textit{Description Construction}} \\
Hierarchical merging & LLaMA-3 & 42.10 & 83.09 \\
 & GPT-3.5 & 39.85 & 80.00  \\
 &  GPT-4 & 45.43 & 85.24 \\
 \hdashline
 \addlinespace[0.05cm]
Incremental updating & LLaMA-3 & 43.82 & 83.21 \\
 & GPT-3.5 & 41.06 & 81.63  \\
 & GPT-4 & 47.02  & 86.47  \\
 \hdashline
 \addlinespace[0.05cm]
Human Description & LLaMA-3 & 52.51 & 87.28 \\
 & GPT-3.5 & 52.04 & 86.33 \\
 & GPT-4 & 55.17 & 90.23 \\
\rowcolor[gray]{0.95} \multicolumn{4}{c}{\textit{Memory Retrieval}}\\
BM25 & GPT-4 & 26.08  & 75.88  \\
Embedding & GPT-4 & 35.66 & 78.24  \\
\rowcolor[gray]{0.95} \multicolumn{4}{c}
{\textit{Description \& Memory}}\\
Direct concatenation & LLaMA-3 & 57.02 & 92.04 \\
& Mixtral & 58.56 & 91.75 \\
& Claude-3.5 & 62.85 & 96.45 \\
& Gemini-1.5-pro & 57.16 & 91.38 \\
& GPT-3.5 & 55.62 & 90.39  \\
& GPT-4 & 62.92 & 95.46  \\
\hdashline
\addlinespace[0.05cm]
\method & LLaMA-3 & 63.72 & 95.93 \\
& Mixtral & 65.02 & 92.05 \\
& Claude-3.5 & 67.13 & \textbf{96.90} \\
& Gemini-1.5-pro & 63.94 & 91.39 \\
& GPT-3.5 & 61.62 & 90.95  \\
& GPT-4 & \textbf{67.95} & 96.87  \\
\bottomrule
\end{tabular}
\caption{Results of different LLMs on \dataset. \textbf{ACC} refers to the decision accuracy. \textit{\textbf{+motivation}} refers to the results after providing the motivations behind  character decisions, which are extracted from expert analyses by GPT-4.}
\label{tab:main_result}
\end{table}

Because our inputs generally exceed 100k, it is difficult for LLMs to handle them directly. Therefore, our approach is divided into two steps: 
1) \textbf{Character Profile Construction}, which includes the character’s description and memories; 
2) \textbf{Reasoning for Decisions}, where different LLMs use the constructed profile to answer the questions.

\subsection{Character Profile Construction}
As shown in Figure \ref{fig:main}, the character profile consists of two parts. The first part is the character's \textbf{description}, including their personality, experiences, hobbies, etc. The second part is the character's \textbf{memories}, specific segments from the preceding text. Below, I will detail the methods for constructing these two parts:

\paragraph{Description Construction}
We adopt two automatic methods to construct character descriptions:
(1) Hierarchical merging~\cite{wu2021recursively}: Books are divided into chunks that fit within the LLM context window. The LLM summarizes each chunk, then merges and summarizes adjacent summarized chunks iteratively to produce the final description.
(2) Incremental updating~\cite{chang2023booookscore}: Books are divided into chunks and summarized sequentially, and the description is updated and refined incrementally by concatenating summarized chunks.
The summarization model for both automated methods is GPT-3.5.
Additionally, using the (3) expert-written descriptions from \textit{Supersummary}, we employ GPT-4 to identify the positions of the decision points and truncate the text, providing only the data before these points.
All descriptions are kept within 5k tokens, the maximum for human-written descriptions.

\paragraph{Memory Retrieval}
We use two memory retrieval methods:
(1) BM25~\cite{robertson2009probabilistic}: Scores documents based on term relevance and length, optimizing retrieval using term frequency and distribution.
(2) Embedding-based retrieval: Uses dense vectors representing documents and queries to assess semantic similarity through vector distance. For the embedding model, we use OpenAI's text-embedding-ada-002\cite{neelakantan2022text} model.

\begin{figure}[t]
    \centering
    \includegraphics[width=\linewidth]{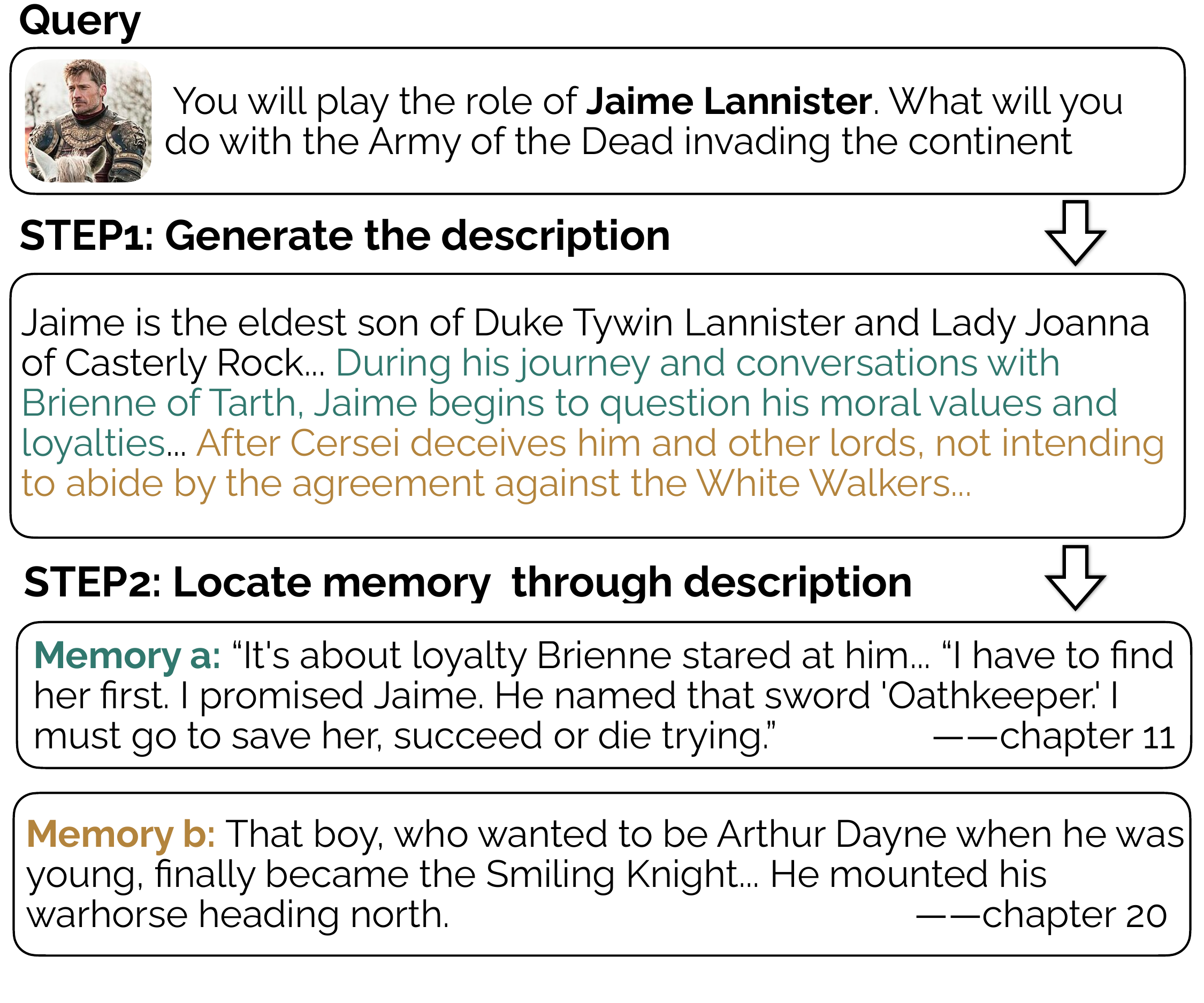}
    \caption{An overview of \method, a two-step scenario-specific character profile building approach.}
    \label{fig:drama}
\end{figure}

\paragraph{Description \& Memory}
Using only Description or Memory alone may lead to information loss~\citep{wang2023does}. Therefore, we also experiment by combining the results of both methods to form the character's profile. We adopt two methods:
(1) Direct concatenation: This method concatenates the results from both approaches by prompting the model to role-play the corresponding character. By default, it uses the results from Human Description and Embedding retrieval.
(2) \method: To better utilize the information in the Description, we propose CHARacter MAPping Profile Synthesis (\method), constructing a more scenario-specific profile in two steps. As shown in Figure \ref{fig:drama}, first, after obtaining the description, we input it along with the question into the model, asking it to locate the plot in the Description relevant to the current scene based on the question. Second, we use these episodes as queries to retrieve related memories and then input them into the inference model and the description. This leverages the overall character storyline in the description, thereby better retrieving related memories.

\subsection{Reasoning for Decisions}
After compressing the original input $x$ into a character profile, we feed it into the LLMs. For methods using only description or memory, we use GPT-3.5, GPT-4, and LLaMA-3\cite{meta2024llama3}. For methods using both, we also include Claude-3\cite{anthropic2024claude3}, Gemini\cite{geminiteam2024gemini}, and Mixtral~\cite{jiang2024mixtral}. For all these models, we adopt the official instruction formats where available \footnote{The versions in this paper are \texttt{gpt-3.5-turbo-1106}, \texttt{gpt-4-1106-preview, Llama-3-70B-Instruct, Claude}
\texttt{-3.5-Sonnet, Gemini-1.5-pro} and \texttt{Mixtral-8x7B-v0.1} respectively.}.




\section{Analysis}
\label{sec:analysis}

\begin{table}[t]
\centering
\small
\begin{tabular}{cccc}
\toprule
& \textbf{Raw text} & \textbf{Concat.} & \textbf{\method} \\
\midrule
GPT-4 & - & 63.05 & 68.10 \\
human & 92.01 & 66.82 & 74.78 \\ \bottomrule
\end{tabular}
\caption{Results of the human evaluation. Concat. refers to the direct concatenation of Description and Memory.}
\label{tab:human}
\end{table}

In the experiments, we wish to answer two research questions:
\begin{inparaenum}[\it RQ1)]
    \item Can LLMs make decisions based on historical data?
    \item What influences the decision-making of LLMs?
\end{inparaenum}

\subsection{Can LLMs Make Decisions Based on Historical Data?}

\paragraph{Analysis of Model Results}

Table \ref{tab:main_result} presents the accuracy results of different RPLA methods on the \dataset. Additionally, we evaluate the results when the model is provided with gold motivation, and several observations can be made:
First, the method that uses both Description and Memory surpasses the one that uses only one, suggesting that both holistic and detailed data of key characters are essential in final decision-making.
Second, when gold motivation is provided, the accuracy consistently exceeds 80\%, indicating the rationality of these motivations in the data.
Third, the performance gap among different LLMs is not significant while reasoning the answer. This indicates that the main factor for the result is the generated profile rather than reasoning ability.
Last, \method outperforms the method that directly concatenates Description and Memory by 5.03\%, proving its effectiveness. This scenario-specific profile better assists RPLA in decision-making.

\paragraph{Humans are Good Decision-makers}
We invite three native English-speaking university students to take a test in which we select six novels they have never heard of before. 
Each novel has between 3 to 5 characters and their corresponding multiple-choice questions. 
We provide each person with three data sets for each key character in two books: the full original text before the decision point, direct concatenation Description and Memory result, and the result from \method.
As shown in Table \ref{tab:human}, compared to direct concatenation, the \method results are easier for humans to understand. 
Additionally, humans slightly outperform GPT-4 in reasoning answers based on the profiles, indicating that humans can understand subtle character decisions better than models. 
When given the raw text, humans can achieve an accuracy rate of 92.01\%, suggesting there is still significant room for improvement in RPLA methods.

\begin{figure}[t]
    \centering
    \includegraphics[width=0.9\linewidth,height=.81\linewidth]{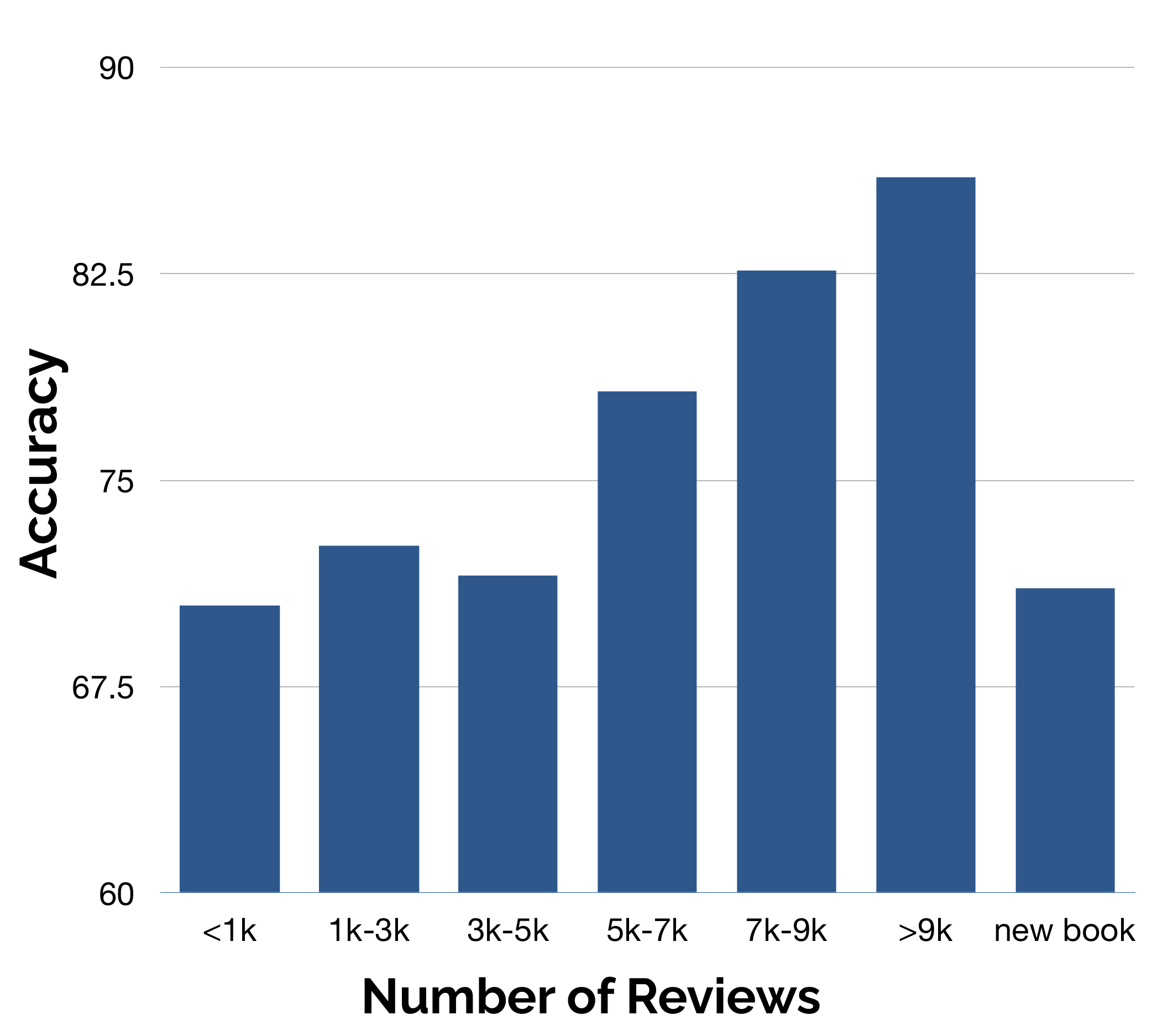}
    \caption{The impact of the number of book reviews on accuracy in \dataset, with \textit{new books} being those not present in the training corpus of LLMs.}
    \label{fig:dataleak}
\end{figure}

\paragraph{Mitigation and Analysis of Data Leakage}

Data leakage is a significant challenge since our data might appear in the model's pre-training corpus. During the data collection phases in section \ref{sec:dataset_construction}, we adopt various preventive measures. For evaluation, we employ an entity replacement strategy, substituting character names, locations, and other entities with placeholders. 
We believe data leakage relates to the amount of relevant corpus used during LLM pre-training, with more popular books having more related corpus.
To verify this, we use the number of reviews on the book review website\footnote{https://www.douban.com/} to indicate a book's popularity and evaluate the results of books with different review counts on \dataset. 
We use \method to build profiles and GPT-4 as the role-playing model, sampling thirty books with different numbers of reviews, including thirty books not in the LLMs' corpus (published after November 6 for gpt-4-1106-preview).
As shown in Figure \ref{fig:dataleak}, the model's accuracy significantly improves when the number of reviews exceeds 5,000. In contrast, books with fewer than 5,000 reviews show slight fluctuation and results similar to those not in the LLMs' corpus. 
Therefore, it can be considered that for books with a low number of reviews, data leakage has little impact on \method. 
In section \ref{sec:dataset_construction}, we use 5,000 reviews as a threshold to filter the books.

\begin{table}[t]
    \centering
    \small
    \begin{tabular}{llcc}
        \toprule
        \textbf{LLMs} & \textbf{Method} & \textbf{Accuracy} \\
        \midrule
        Claude-3.5 & long-context & 64.83 \\
        Claude-3.5 & \method & 67.13 \\
        Genmini-1.5 Pro & long-context & 61.14 \\
        Genmini-1.5 Pro & \method & 63.94 \\
        \bottomrule
    \end{tabular}
    \caption{The results of using long-context models for \dataset.}
    \label{tab:longcontext}
\end{table}
\paragraph{Analysis of Long-Context LLMs}
Long context is an essential feature of \dataset, and directly using long-context models for role-playing is an exciting topic. 
Making decisions based on extensive context tests a model’s ability to understand global data and reason from a character's perspective. 
We evaluate two long-context models: Claude3-sonnect and kimi-chat. 
As shown in Table \ref{tab:longcontext}, although the performance of long-context models is not as strong as \method, they still demonstrate potential in role-playing. \dataset, as a task requiring multiple reasoning points and an overall understanding of the context, can also serve as a vital benchmark for evaluating long-context models.

\subsection{What Influences the Decision-making of LLMs?}
\label{sec:influence}

\begin{figure}[t]
    \centering
    \includegraphics[width=.85\linewidth,height=\linewidth]{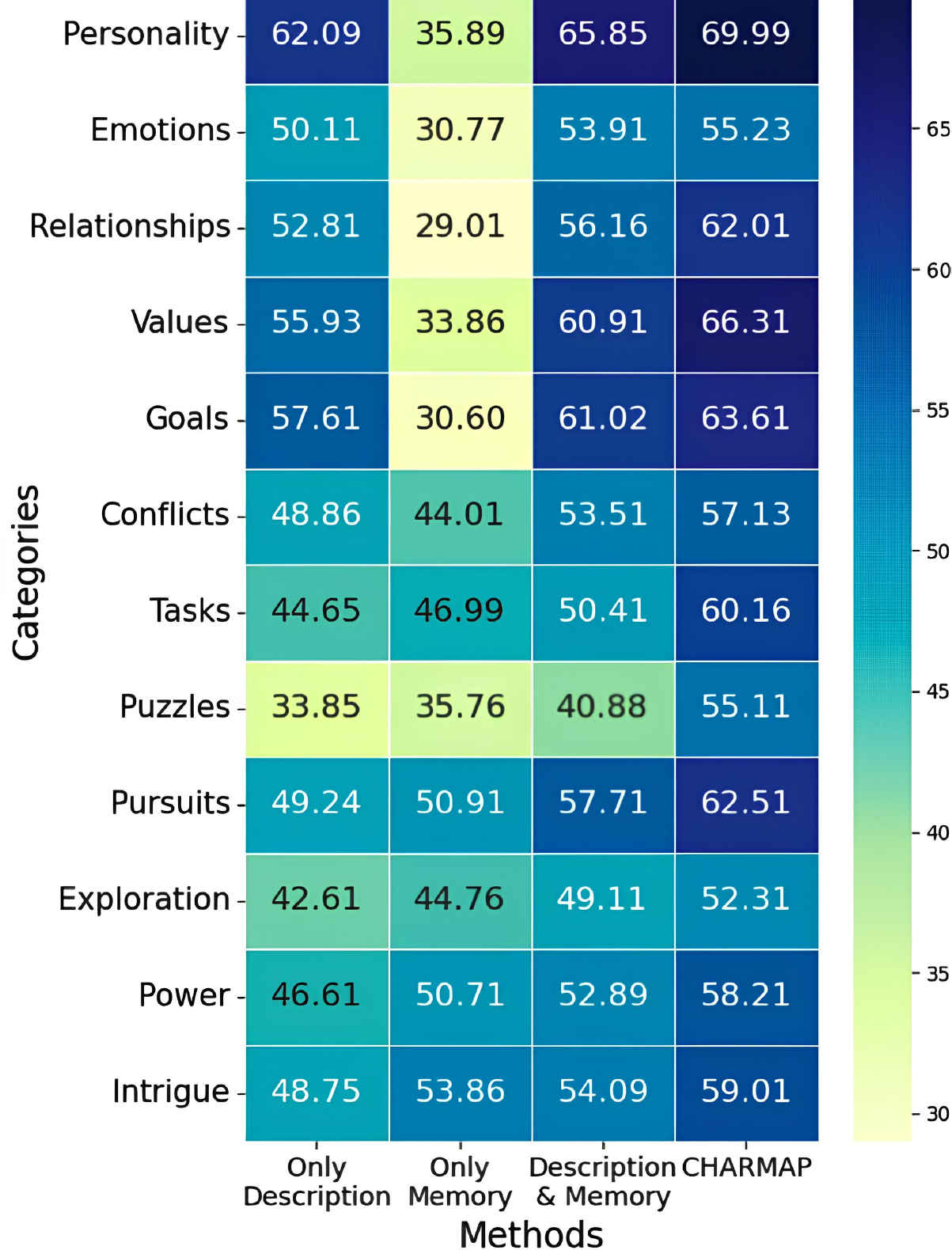}
    \caption{Heatmap of the impact of motivation types on the results. The results are predicted from the Incremental updating, the embedding-retrieved memory, the direct concatenation of both, and \method. The role-playing model uses GPT-4.}
    \label{fig:heat}
\end{figure}

\paragraph{The Impact of Motivation Types}
In line with the motivation types presented in Section \ref{sec:dataset_analysis}, we examine how different types of motivation influence characters' decision-making.
For profiles, we evaluate four methods: the Incremental updating, the embedding-retrieved memory,
the direct concatenation of both, and \method.
For reasoning, we use GPT-4 uniformly. 
The results are shown in Figure \ref{fig:heat}. 
We find that tasks requiring coherent reasoning, such as puzzles and mysteries, are not well answered for all methods.
This might be because these questions need multi-step reasoning and details from various memories.
Moreover, plot-driven questions have lower accuracy when descriptions are used only for the profile. 
Conversely, character-driven questions are challenging to answer when relying only on memories. We believe this is because character summaries in descriptions better capture the overall essence of the characters, while memories provide direct access to relevant events.

\begin{figure}[t]
    \centering
    \includegraphics[width=\linewidth]{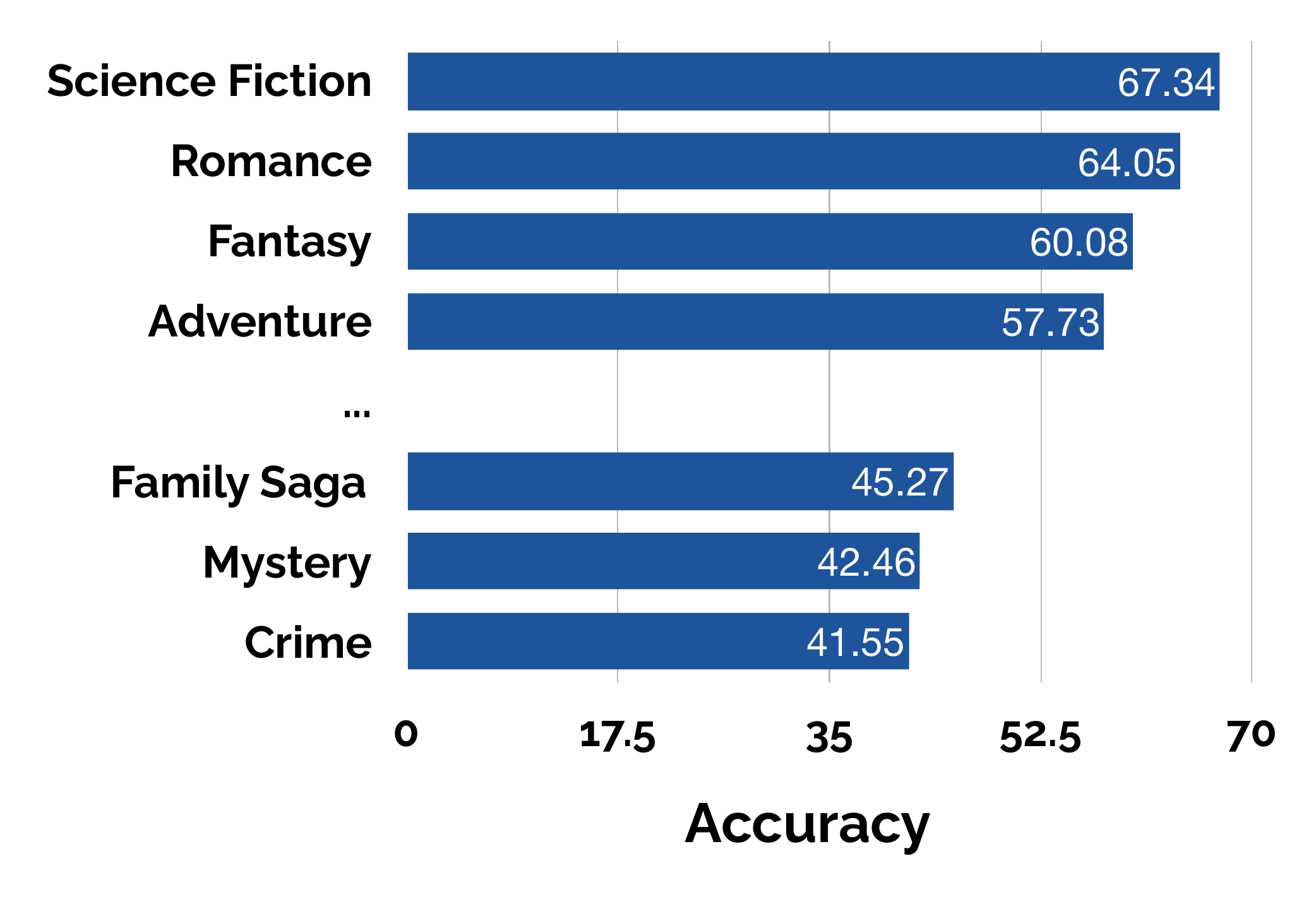}
    \caption{The result of the impact of different novel genres on accuracy.}
    \label{fig:genre}
\end{figure}

\paragraph{The Impact of Novel Genres}
We use the genre tags from novels on the website to analyze the accuracy of character selection across different genres.
We conduct experiments on the the direct concatenation of description and memories, and the role-playing model using GPT-4.
As depicted in Figure \ref{fig:genre}, the accuracy of science fiction, fantasy novels, and romance novels is quite high. 
This could be because the characters in these novels are often stylized or have fixed creative patterns and archetypes. 
In contrast, crime and mystery novels perform poorly, which might be because they involve complex logical chains, and characters in these novels frequently take abnormal actions. 
Further details about each genre and the complete table can be found in Appendix \ref{appendix:category_1}.

\begin{figure}[t]
    \centering
    \includegraphics[width=.8\linewidth,height=.75\linewidth]{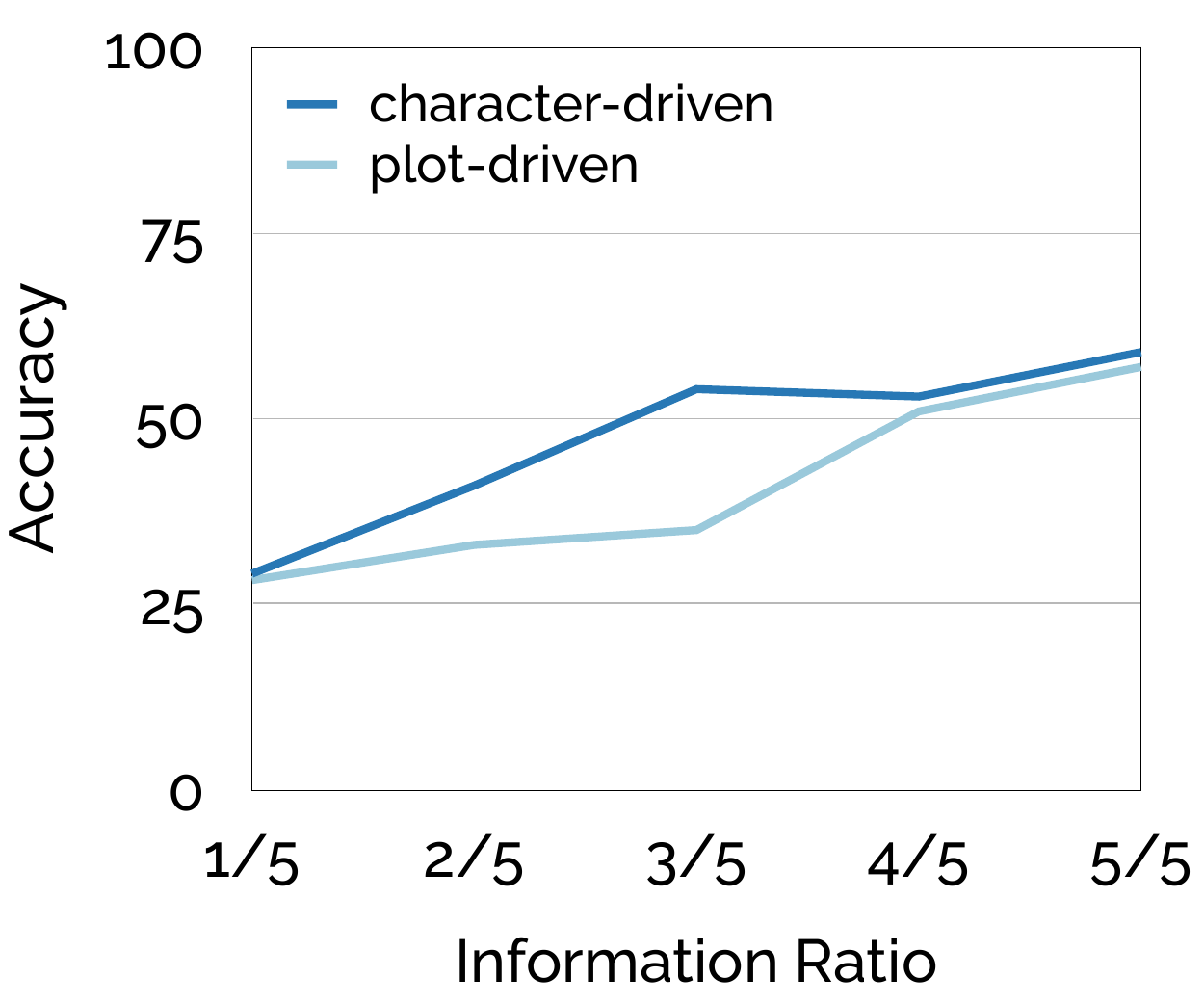}
    \caption{Analysis of whether character selection will change. The x-axis represents the input length relative to the point truncation.}
    \label{fig:timeline_3}
\end{figure}

\paragraph{The Impact of Temporal Data}
If faced with the decisions of years past at this moment, would you make the same choices? We conduct a study on this matter. 
Specifically, we randomly sample 40 characters, half character-driven, and half plot-driven. 
We split the content preceding the decision points into five equal sections and used these various content lengths as input.
We conduct experiments on the combination of human description + embedding-retrieved memories, and the role-playing model is GPT-4.
As shown in Figure \ref{fig:timeline_3}, in the early stages, the accuracy of most characters' decisions is close to random (25\%), potentially due to insufficient information. 
As more information becomes available, the characters' decisions tend to be closer to the correct choice. 
For character-driven decisions, accuracy tends to be stable. For plot-driven, the accuracy rate may change abruptly. 
This could be due to the relatively stable characteristics of a character, while some sudden events may greatly influence the final choices of the character.



\section{Conclusion}
\label{sec:conclusion}

In this work, we propose the first task to evaluate the decision-making of RPLAs, testing whether LLMs can accurately reconstruct decisions using historical data. We construct \dataset, which includes 1,462 characters from 388 books and their life choices. Extensive experiments on \dataset demonstrate the promising performance of RPLAs in decision simulation. Additionally, we propose \method, which uses persona-based memory retrieval to enhance decision-making. We hope this work provides better evaluation benchmarks for RPLAs and directs the future development of personal LLM assistants.


\clearpage
\section*{Limitations}
\label{sec:lim}
In this paper, we primarily investigate whether fictional characters can recreate their choices within a book. Although we have controlled the quality of the novels, there may still be issues with the plot and characters since the author designed the storyline, which can result in illogical choices within the book. Furthermore, as our research focuses mainly on fictional characters, there is a certain gap compared to real-world humans. For example, the author fictionalizes some story backgrounds, which may impact the model's generation results.

\section*{Ethics Statement}
\label{sec:ethic}
\paragraph{Use of Human Annotations}
Our institution recruits annotators to implement the annotations of motivation recognition dataset construction. 
We ensure the privacy rights of the annotators are respected during the annotation process.
The annotators receive compensation exceeding the local minimum wage and have consented to use motivation recognition data they process for research purposes. Appendix~\ref{sec:annotation} provides further details on the annotations.

\paragraph{Risks}
The \dataset dataset is sourced from novels and analysis data written by human literary experts. However, we cannot guarantee that this data is entirely free from toxic or discriminatory language. Additionally, our data is generated by GPT-4, which may introduce some inherent biases and hallucinations of the model.

\bibliography{custom}

\begin{thebibliography}{36}
\expandafter\ifx\csname natexlab\endcsname\relax\def\natexlab#1{#1}\fi

\bibitem[{Anthropic(2024)}]{anthropic2024claude3}
Anthropic. 2024.
\newblock \href {https://www-cdn.anthropic.com/de8ba9b01c9ab7cbabf5c33b80b7bbc618857627/Model_Card_Claude_3.pdf} {The claude 3 model family: Opus, sonnet, haiku}.

\bibitem[{Brahman et~al.(2021)Brahman, Huang, Tafjord, Zhao, Sachan, and Chaturvedi}]{brahman2021let}
Faeze Brahman, Meng Huang, Oyvind Tafjord, Chao Zhao, Mrinmaya Sachan, and Snigdha Chaturvedi. 2021.
\newblock " let your characters tell their story": A dataset for character-centric narrative understanding.
\newblock \emph{arXiv preprint arXiv:2109.05438}.

\bibitem[{Chang et~al.(2023)Chang, Lo, Goyal, and Iyyer}]{chang2023booookscore}
Yapei Chang, Kyle Lo, Tanya Goyal, and Mohit Iyyer. 2023.
\newblock Booookscore: A systematic exploration of book-length summarization in the era of llms.
\newblock \emph{arXiv preprint arXiv:2310.00785}.

\bibitem[{Chen et~al.(2024)Chen, Wang, Xu, Yuan, Zhang, Shi, Xie, Li, Yang, Zhu, Chen, Li, Chen, Hu, Wu, Ren, Fu, and Xiao}]{chen2024persona}
Jiangjie Chen, Xintao Wang, Rui Xu, Siyu Yuan, Yikai Zhang, Wei Shi, Jian Xie, Shuang Li, Ruihan Yang, Tinghui Zhu, Aili Chen, Nianqi Li, Lida Chen, Caiyu Hu, Siye Wu, Scott Ren, Ziquan Fu, and Yanghua Xiao. 2024.
\newblock \href {http://arxiv.org/abs/2404.18231} {From persona to personalization: A survey on role-playing language agents}.

\bibitem[{Dodge and Kitchin(2007)}]{dodge2007outlines}
Martin Dodge and Rob Kitchin. 2007.
\newblock ‘outlines of a world coming into existence’: pervasive computing and the ethics of forgetting.
\newblock \emph{Environment and planning B: planning and design}, 34(3):431--445.

\bibitem[{Gao et~al.(2023)Gao, Lian, Zhou, Fu, and Wang}]{gao2023livechat}
Jingsheng Gao, Yixin Lian, Ziyi Zhou, Yuzhuo Fu, and Baoyuan Wang. 2023.
\newblock Livechat: A large-scale personalized dialogue dataset automatically constructed from live streaming.
\newblock \emph{arXiv preprint arXiv:2306.08401}.

\bibitem[{Gao et~al.(2024)Gao, Xiong, Gao, Jia, Pan, Bi, Dai, Sun, Guo, Wang, and Wang}]{gao2024retrievalaugmented}
Yunfan Gao, Yun Xiong, Xinyu Gao, Kangxiang Jia, Jinliu Pan, Yuxi Bi, Yi~Dai, Jiawei Sun, Qianyu Guo, Meng Wang, and Haofen Wang. 2024.
\newblock \href {http://arxiv.org/abs/2312.10997} {Retrieval-augmented generation for large language models: A survey}.

\bibitem[{Gurrin et~al.(2014)Gurrin, Smeaton, Doherty et~al.}]{gurrin2014lifelogging}
Cathal Gurrin, Alan~F Smeaton, Aiden~R Doherty, et~al. 2014.
\newblock Lifelogging: Personal big data.
\newblock \emph{Foundations and Trends{\textregistered} in information retrieval}, 8(1):1--125.

\bibitem[{Hoy(2018)}]{hoy2018alexa}
Matthew~B Hoy. 2018.
\newblock Alexa, siri, cortana, and more: an introduction to voice assistants.
\newblock \emph{Medical reference services quarterly}, 37(1):81--88.

\bibitem[{Jaiswal et~al.(2020)Jaiswal, Raju, and Deb}]{jaiswal2020facial}
Akriti Jaiswal, A~Krishnama Raju, and Suman Deb. 2020.
\newblock Facial emotion detection using deep learning.
\newblock In \emph{2020 international conference for emerging technology (INCET)}, pages 1--5. IEEE.

\bibitem[{Jiang et~al.(2024)Jiang, Sablayrolles, Roux, Mensch, Savary, Bamford, Chaplot, de~las Casas, Hanna, Bressand, Lengyel, Bour, Lample, Lavaud, Saulnier, Lachaux, Stock, Subramanian, Yang, Antoniak, Scao, Gervet, Lavril, Wang, Lacroix, and Sayed}]{jiang2024mixtral}
Albert~Q. Jiang, Alexandre Sablayrolles, Antoine Roux, Arthur Mensch, Blanche Savary, Chris Bamford, Devendra~Singh Chaplot, Diego de~las Casas, Emma~Bou Hanna, Florian Bressand, Gianna Lengyel, Guillaume Bour, Guillaume Lample, Lélio~Renard Lavaud, Lucile Saulnier, Marie-Anne Lachaux, Pierre Stock, Sandeep Subramanian, Sophia Yang, Szymon Antoniak, Teven~Le Scao, Théophile Gervet, Thibaut Lavril, Thomas Wang, Timothée Lacroix, and William~El Sayed. 2024.
\newblock \href {http://arxiv.org/abs/2401.04088} {Mixtral of experts}.

\bibitem[{Kaplan and Haenlein(2019)}]{kaplan2019siri}
Andreas Kaplan and Michael Haenlein. 2019.
\newblock Siri, siri, in my hand: Who’s the fairest in the land? on the interpretations, illustrations, and implications of artificial intelligence.
\newblock \emph{Business horizons}, 62(1):15--25.

\bibitem[{Li et~al.(2023)Li, Leng, Yan, Shen, Wang, MI, Fei, Feng, Yan, Wang et~al.}]{li2023chatharuhi}
Cheng Li, Ziang Leng, Chenxi Yan, Junyi Shen, Hao Wang, Weishi MI, Yaying Fei, Xiaoyang Feng, Song Yan, HaoSheng Wang, et~al. 2023.
\newblock Chatharuhi: Reviving anime character in reality via large language model.
\newblock \emph{arXiv preprint arXiv:2308.09597}.

\bibitem[{Li et~al.(2018)Li, Yang, Guo, Chen, Agarwal, and Hong}]{li2018automated}
Yuanchun Li, Ziyue Yang, Yao Guo, Xiangqun Chen, Yuvraj Agarwal, and Jason~I Hong. 2018.
\newblock Automated extraction of personal knowledge from smartphone push notifications.
\newblock In \emph{2018 IEEE International Conference on Big Data (Big Data)}, pages 733--742. IEEE.

\bibitem[{Majumder et~al.(2017)Majumder, Poria, Gelbukh, and Cambria}]{majumder2017deep}
Navonil Majumder, Soujanya Poria, Alexander Gelbukh, and Erik Cambria. 2017.
\newblock Deep learning-based document modeling for personality detection from text.
\newblock \emph{IEEE Intelligent Systems}, 32(2):74--79.

\bibitem[{Neelakantan et~al.(2022)Neelakantan, Xu, Puri, Radford, Han, Tworek, Yuan, Tezak, Kim, Hallacy, Heidecke, Shyam, Power, Nekoul, Sastry, Krueger, Schnurr, Such, Hsu, Thompson, Khan, Sherbakov, Jang, Welinder, and Weng}]{neelakantan2022text}
Arvind Neelakantan, Tao Xu, Raul Puri, Alec Radford, Jesse~Michael Han, Jerry Tworek, Qiming Yuan, Nikolas Tezak, Jong~Wook Kim, Chris Hallacy, Johannes Heidecke, Pranav Shyam, Boris Power, Tyna~Eloundou Nekoul, Girish Sastry, Gretchen Krueger, David Schnurr, Felipe~Petroski Such, Kenny Hsu, Madeleine Thompson, Tabarak Khan, Toki Sherbakov, Joanne Jang, Peter Welinder, and Lilian Weng. 2022.
\newblock \href {http://arxiv.org/abs/2201.10005} {Text and code embeddings by contrastive pre-training}.

\bibitem[{OpenAI(2023)}]{openai2023gpt4}
OpenAI. 2023.
\newblock \href {http://arxiv.org/abs/2303.08774} {Gpt-4 technical report}.

\bibitem[{Park et~al.(2023)Park, O'Brien, Cai, Morris, Liang, and Bernstein}]{park2023generative}
Joon~Sung Park, Joseph~C. O'Brien, Carrie~J. Cai, Meredith~Ringel Morris, Percy Liang, and Michael~S. Bernstein. 2023.
\newblock \href {http://arxiv.org/abs/2304.03442} {Generative agents: Interactive simulacra of human behavior}.

\bibitem[{Robertson et~al.(2009)Robertson, Zaragoza et~al.}]{robertson2009probabilistic}
Stephen Robertson, Hugo Zaragoza, et~al. 2009.
\newblock The probabilistic relevance framework: Bm25 and beyond.
\newblock \emph{Foundations and Trends{\textregistered} in Information Retrieval}, 3(4):333--389.

\bibitem[{Salemi et~al.(2024)Salemi, Mysore, Bendersky, and Zamani}]{salemi2024lamp}
Alireza Salemi, Sheshera Mysore, Michael Bendersky, and Hamed Zamani. 2024.
\newblock \href {http://arxiv.org/abs/2304.11406} {Lamp: When large language models meet personalization}.

\bibitem[{Shao et~al.(2023)Shao, Li, Dai, and Qiu}]{shao2023character}
Yunfan Shao, Linyang Li, Junqi Dai, and Xipeng Qiu. 2023.
\newblock Character-llm: A trainable agent for role-playing.
\newblock \emph{arXiv preprint arXiv:2310.10158}.

\bibitem[{Silk(2002)}]{silk2002aristophanes}
Michael~Stephen Silk. 2002.
\newblock \emph{Aristophanes and the Definition of Comedy}.
\newblock Oxford University Press, USA.

\bibitem[{Sommerstein(2013)}]{sommerstein2013aristophanes}
Alan Sommerstein. 2013.
\newblock Aristophanes.
\newblock \emph{The Encyclopedia of Ancient History}.

\bibitem[{{\v{S}}tajner and Yenikent(2020)}]{vstajner2020survey}
Sanja {\v{S}}tajner and Seren Yenikent. 2020.
\newblock A survey of automatic personality detection from texts.
\newblock In \emph{Proceedings of the 28th international conference on computational linguistics}, pages 6284--6295.

\bibitem[{Tang et~al.(2019)Tang, Wang, Yang, and Song}]{tang2019akupm}
Xiaoli Tang, Tengyun Wang, Haizhi Yang, and Hengjie Song. 2019.
\newblock Akupm: Attention-enhanced knowledge-aware user preference model for recommendation.
\newblock In \emph{Proceedings of the 25th ACM SIGKDD international conference on knowledge discovery \& data mining}, pages 1891--1899.

\bibitem[{Team(2024{\natexlab{a}})}]{geminiteam2024gemini}
Gemini Team. 2024{\natexlab{a}}.
\newblock \href {http://arxiv.org/abs/2312.11805} {Gemini: A family of highly capable multimodal models}.

\bibitem[{Team(2024{\natexlab{b}})}]{meta2024llama3}
Meta~LLaMA Team. 2024{\natexlab{b}}.
\newblock Introducing meta llama 3: The most capable openly available llm to date.
\newblock \url{https://ai.meta.com/blog/meta-llama-3/}.
\newblock Accessed: 2023-10-03.

\bibitem[{Touvron et~al.(2023)Touvron, Martin, Stone, Albert, and Almahairi}]{touvron2023llama}
Hugo Touvron, Louis Martin, Kevin Stone, Peter Albert, and Amjad Almahairi. 2023.
\newblock \href {http://arxiv.org/abs/2307.09288} {Llama 2: Open foundation and fine-tuned chat models}.

\bibitem[{Wang et~al.(2024)Wang, Xiao, tse Huang, Yuan, Xu, Guo, Tu, Fei, Leng, Wang, Chen, Li, and Xiao}]{wang2023does}
Xintao Wang, Yunze Xiao, Jen tse Huang, Siyu Yuan, Rui Xu, Haoran Guo, Quan Tu, Yaying Fei, Ziang Leng, Wei Wang, Jiangjie Chen, Cheng Li, and Yanghua Xiao. 2024.
\newblock \href {http://arxiv.org/abs/2310.17976} {Incharacter: Evaluating personality fidelity in role-playing agents through psychological interviews}.

\bibitem[{Wang et~al.(2023)Wang, Peng, Que, Liu, Zhou, Wu, Guo, Gan, Ni, Zhang, Zhang, Ouyang, Xu, Chen, Fu, and Peng}]{wang2023rolellm}
Zekun~Moore Wang, Zhongyuan Peng, Haoran Que, Jiaheng Liu, Wangchunshu Zhou, Yuhan Wu, Hongcheng Guo, Ruitong Gan, Zehao Ni, Man Zhang, Zhaoxiang Zhang, Wanli Ouyang, Ke~Xu, Wenhu Chen, Jie Fu, and Junran Peng. 2023.
\newblock \href {http://arxiv.org/abs/2310.00746} {Rolellm: Benchmarking, eliciting, and enhancing role-playing abilities of large language models}.

\bibitem[{Wu et~al.(2021)Wu, Ouyang, Ziegler, Stiennon, Lowe, Leike, and Christiano}]{wu2021recursively}
Jeff Wu, Long Ouyang, Daniel~M. Ziegler, Nisan Stiennon, Ryan Lowe, Jan Leike, and Paul Christiano. 2021.
\newblock \href {http://arxiv.org/abs/2109.10862} {Recursively summarizing books with human feedback}.

\bibitem[{Xie et~al.(2024)Xie, Chen, Jia, Ye, Shu, Bibi, Hu, Torr, Ghanem, and Li}]{xie2024can}
Chengxing Xie, Canyu Chen, Feiran Jia, Ziyu Ye, Kai Shu, Adel Bibi, Ziniu Hu, Philip Torr, Bernard Ghanem, and Guohao Li. 2024.
\newblock Can large language model agents simulate human trust behaviors?
\newblock \emph{arXiv preprint arXiv:2402.04559}.

\bibitem[{Xu et~al.(2022)Xu, Li, Wang, Yang, Wang, and Xiao}]{Xu_2022}
Chen Xu, Piji Li, Wei Wang, Haoran Yang, Siyun Wang, and Chuangbai Xiao. 2022.
\newblock \href {https://doi.org/10.1145/3477495.3531957} {Cosplay: Concept set guided personalized dialogue generation across both party personas}.
\newblock In \emph{Proceedings of the 45th International ACM SIGIR Conference on Research and Development in Information Retrieval}, SIGIR ’22. ACM.

\bibitem[{Yu et~al.(2022)Yu, Sang, Pu, Wei, Wang, Li, Yu, and Zhou}]{yu2022few}
Mo~Yu, Yisi Sang, Kangsheng Pu, Zekai Wei, Han Wang, Jing Li, Yue Yu, and Jie Zhou. 2022.
\newblock Few-shot character understanding in movies as an assessment to meta-learning of theory-of-mind.
\newblock \emph{arXiv preprint arXiv:2211.04684}.

\bibitem[{Zad et~al.(2021)Zad, Heidari, James~Jr, and Uzuner}]{zad2021emotion}
Samira Zad, Maryam Heidari, H~James~Jr, and Ozlem Uzuner. 2021.
\newblock Emotion detection of textual data: An interdisciplinary survey.
\newblock In \emph{2021 IEEE World AI IoT Congress (AIIoT)}, pages 0255--0261. IEEE.

\bibitem[{Zhou et~al.(2023)Zhou, Chen, Wan, Wen, Song, Yu, Huang, Peng, Yang, Xiao et~al.}]{zhou2023characterglm}
Jinfeng Zhou, Zhuang Chen, Dazhen Wan, Bosi Wen, Yi~Song, Jifan Yu, Yongkang Huang, Libiao Peng, Jiaming Yang, Xiyao Xiao, et~al. 2023.
\newblock Characterglm: Customizing chinese conversational ai characters with large language models.
\newblock \emph{arXiv preprint arXiv:2311.16832}.

\end{thebibliography}
\bibliographystyle{acl_natbib}

\clearpage
\appendix
\label{sec:appendix}
\begin{table*}[t]
  \centering
  \small
    \begin{tabularx}{\linewidth}{X}

    \toprule
    
    \textbf{Key Character Description }\\
    
    \toprule
    
    \textbf{John “Jack” Robert Lee} \\

    \midrule

    Jack Lee is the protagonist of the novel. He is a white man who sees himself as intelligent, having done well in school and having a law degree, yet he acknowledges that he did not go to one of the best law schools. At the beginning of the novel, he turns 33. He is disappointed with much of his professional life, having been out of law school for eight years and still “just getting by.” He believes that he has largely failed to “change” the world—which was one of his goals in becoming a lawyer.\\
    Jack changes throughout the course of the novel. At the novel’s start, he recognizes the racist actions of his mother and is glad that segregation is ending. Because of his love of books, he has a vast knowledge of Black history and the hardships that Black people have faced, yet he largely ignores these hardships, as they do not affect him. He largely chooses to go along with the way things are and scolds himself for not being a “risk-taker.”\\
    However, he realizes how very real injustice is for people like Jerome. He sets aside his own fear and faces the danger of representing him. Initially, he dislikes DuBose’s interest in speaking with the press and trying to use Jerome’s trial as anything other than a chance to save Jerome’s life. By contrast, he speaks to the press for the first time at the novel’s conclusion, making an impassioned plea about the importance of coming together as a community.\\
    
    \midrule

    \textbf{Desiree DuBose} \\

    \midrule

    DuBose is a Black lawyer from Chicago. She is extremely intelligent, having gone to college at age 16 and graduated from Yale Law School after six years. She has a vast array of experience working with the NAACP and the Legal Defense Fund to fight against racist legislation, including, mostly recently, the Loving case, which allowed Black and white people to marry. She initially comes to Virginia with the intent of taking over Jerome’s defense from Jack, but after seeing how committed he is to the case, agrees to be his co-counselor.\\
    DuBose contrasts with Jack. She is very different from him on the surface level: She is a woman, Black, has served as a lawyer in dozens of murder trials, and recognizes the importance of Jerome’s trial to the larger picture of civil rights. However, like Jack, she is a dynamic character in that she changes throughout the text. After Jack asks her to stop focusing on mistrials or appeals, she brings her full effort to the courtroom, fighting back against the admission of the murder weapon instead of utilizing it as a chance for appeal. At the novel’s end, she succeeds in Overcoming Personal Bias by putting aside her fear and hesitancy to be involved with Jack.\\

    \midrule

    \textbf{Jerome Washington} \\

    \midrule

    Jerome is a Black man on trial for the murder of a wealthy white couple. He is a veteran of the Vietnam War and is described as “large” and “strong,” standing at 6’5”. He is dedicated to his job working for the Randolphs before their death, riding his bike five miles each way, never missing a day of work. He is also dedicated to his wife, Pearl and three children; he is willing to go to prison for life if it means that Pearl is able to be acquitted, then is willing to accept a plea of five years, despite the overwhelming evidence in his favor.\\
    Jerome is a flat character in the novel, meaning that he doesn’t change. He serves primarily as a plot device and a way to illuminate the systemic racism and injustice of the time. The narrative offers little information about him. Jack and DuBose repeatedly choose not to put him on the stand to speak even in his own defense. This reflects his status in 1968 in the American South: He has little control of his own life, and is at the mercy of the white people around him and the racist system that he lives in.\\

    \midrule

    \textbf{Hilda “Hilly” Lee} \\

    \midrule

    Hilly is Jack’s mother. A homemaker, she cares for her developmentally disabled daughter, Lucy, well into adulthood while blaming herself for Lucy’s disability. Jack sees his mother as “complicated.” Hilly is upset by Martin Luther King Jr.’s death and helps the Black men who work with her husband, yet is vocal about her belief in segregation. DuBose initially thinks of her as a “typical racist” who perpetuates the idea that she is somehow superior to the Black people around her.\\
    Hilly is a dynamic character who changes throughout the text. As the text progresses, she reverts back to who she originally was prior to the events of the novel—a kind, nonracist woman. After Lucy’s death, she gets to know DuBose and invites her into her home, even lending her clothing. She then reveals that she used to love a Black man, but was forced apart from him. Additionally, she was told by a preacher that Lucy’s disability was a punishment from God in return for loving a Black man. These experiences, and the society in which she lived, led to years of acceptance and eventual perpetuation of racism.\\

    \midrule

    \textbf{Howard Pickett} \\

    \midrule

    Pickett is an antagonist, or villain, of the text. He is a wealthy man who owns coal mines. He is interested in the trial because he wants to use it as a talking point for campaigning for George Wallace to win the 1968 presidential election. Throughout the novel, he speaks with the presses and stresses the importance of Jerome being convicted to emphasize that segregation should be legalized again. Whether Pickett believes these ideas or is simply using them to drum up support for his political campaign is unclear; however, he is unapologetic in his racist language. Pickett represents the true problems that need to be addressed in America. He distracts the working class by perpetuating racism and stressing that Black people are the problem. In this way, he masks that the true issue lies with the greed and theft of the wealthy from the working class.\\
    
    \bottomrule
    
    \end{tabularx}
\caption{Data examples of key character descriptions written by literary experts, sourced from the 2024 novel \textit{"A Calamity of Souls."}}
\label{tab:case_1}
\end{table*}
\begin{table*}[t]
  \centering
  \small
    \begin{tabularx}{\linewidth}{X}

    \toprule
    
    \textbf{Chapter Summaries}\\
    
    \toprule
    
    \textbf{Chapter 1} \\

    \midrule

    In Freeman County, Virginia, in 1968, an elderly white couple is dead in their home. The husband is sprawled across the floor, while his wife’s body lays across a chair.\\
    Two white officers—Raymond Leroy and Gene Taliaferro —have a Black man in handcuffs on the floor, referred to as “the only suspect in the room”. Raymond struggles to read him his Miranda Rights off an index card, a new policy recently enacted in the police force. The idea of reading them annoys both Raymond and Gene, who are bothered by the idea of criminals getting representation, especially “those people, who had committed crimes, usually against white folks”.\\
    Gene interrupts Raymond to hit the suspect with his club. He forces the suspect to lie down, then hits him again, then forces him to kneel again. Gene goads the suspect into getting angry by asking him about his wife and family. When the suspect reacts with rage, struggling against his handcuffs, Gene is excited that he can now claim the suspect was “resistin’ arrest” and raises his club to beat him.\\
    
    \midrule

    \textbf{Chapter 2} \\

    \midrule

    John “Jack” Robert Lee is a lawyer from Freeman County, Virginia. He is white, single, and grew up in a working-class home, with a love of books and debate.\\
    He arrives at his parents’ home to celebrate his 33rd birthday. He is greeted by his older sister, Lucy. She is 37. Due to their mother’s exposure to nitrous oxide at the dentist while pregnant, she is developmentally disabled.\\
    
    \midrule

    \textbf{Chapter 3} \\

    \midrule

    Jack’s mother, Hilda “Hilly” Lee,” has always cared for her home and children while her husband works. She harbors some guilt over what happened to Lucy and chose to have Jack and his younger brother, Jefferson, without any pain killers—even aspirin. She refers to Jack as “Robert,” insisting that she would have named him Robert E. Lee if her husband did not have a say.\\
    Despite naming her son after a Confederate general, she still tells Jack how upset she is about the deaths of Martin Luther King Jr. and Robert Kennedy. Jack finds it “bewildering” that she respects Lee while mourning King and Kennedy, who “held views diametrically opposed to all the Confederacy had stood for.”\\
    Hilly tells Jack that Miss Jessup was by earlier looking for him. Miss Jessup is one of the only Black women in the area. She is a housemaid to Ashby, a wealthy, retired lawyer who lives down the street.\\

    \midrule

    \textbf{...} \\

    \midrule

    \textbf{Chapter 93} \\

    \midrule

    Jack flies to Chicago. On the flight there he thinks of his injury, and how the bullet barely missed doing any major damage. He was also saved by the fact that the bullet went through Jerome first; he is saddened that he can’t thank him for saving his life.\\
    He goes to DuBose’s apartment and surprises her. He tells her that he came to Chicago to work with her. She makes it clear that they can only work together, and Jack admits that he cares for her. She tells him that she once lost a man she loved because of the work that they do, and she can’t go through that pain again. She compares it to Jack’s losing Lucy. Jack insists that meeting her was one of the best things that ever happened to him, despite the damage it caused.\\
    When DuBose doesn’t respond, Jack turns to leave. She stops him, informing him that it will “be far tougher” than he thinks. Jack insists that he is now “far tougher” than he thought he would ever be.\\
    
    \bottomrule
    
    \end{tabularx}
\caption{Data examples of chapter summaries written by literary experts, sourced from the 2024 novel \textit{"A Calamity of Souls."}}
\label{tab:case_2}
\end{table*}
\begin{table*}[t]
  \centering
  \small
    \begin{tabularx}{\linewidth}{X}

    \toprule
    
    \textbf{Book Analysis}\\
    
    \toprule
    
    \textbf{Chapters 1-10} \\

    \midrule

    The setting of the novel plays a pivotal role. Set in the South in 1968, the country is on the verge of moving forward with the end of Jim Crow and segregation, while people throughout the South fight against this. The novel discusses the death of Martin Luther King Jr. as well as the impending presidential election as important moments in history that will decide the future.\\
    Jack battles with Overcoming Personal Bias. He has grown up in the South with a mother who encourages segregation, and up until the novel’s events, has not fought against racism or done something meaningful with his law degree. He struggles with his lack of action, and contemplates whether the danger that will surround the trial is worth it. He recognizes his own bias, and doesn’t yet see racism as an important enough cause to risk his safety. However, when he is confronted with violence on taking Jerome’s case, it has the opposite of its intended effect: Instead of discouraging him from defending Jerome, the violence shows him that he has ignored the problem of racism for too long.\\
    Jack’s mother, Hilly, represents a vast number of white people throughout the South. She is complex: She helps Black people in need but also adamantly believes in segregation. This reflects the beliefs that many people hold throughout the novel. Although she does not believe herself to be racist and feels sympathy for Black people, she still exhibits racist biases and does not want to go against the status quo. Like Jack, she battles with her own personal bias and reflects on whether change is truly needed.\\
    This section begins to examine Racial Injustice and the Legal System. The text reveals the limits that the legal system has when placed in the hands of racist people. Despite what the law says, people continue to perpetuate racism, both directly and indirectly. This reveals that the law struggles without the support of its people.\\
    David Baldacci raises the stakes surrounding the trial to build suspense. For example, after registering himself as Jerome’s lawyer, Jack receives a frightening phone call. The call reinforces his belief that he is doing the right thing using his legal skills, and that he is fighting back against injustice. It foreshadows the danger that will surround the case for both Jack and the people in his life.\\
    
    \midrule

    \textbf{...} \\

    \midrule

    \textbf{Chapters 76-93} \\

    \midrule
    
    This section continues to examine The Importance of Family and Community Support. Like with Lucy’s death, the end of the trial and Jerome’s murder act as catalysts, where people of different races come together. For instance, Jerome’s funeral is attended by many white people who did not even know him. Additionally, Jack notes that people agree with his speech after the trial. In this way, the novel implies hope for the future, and suggests that racial injustice can be overcome.\\
    This section continues to examine Racial Injustice and the Legal System. Jack and DuBose prove several things throughout the trial: Several of the witnesses for the prosecution were pressured into giving false statements, Pearl could not have helped with the crime, and Jerome could not have committed the murder due to his injury. Despite all of this, they are still not able to get the case thrown out and are forced to consider the best plea deal Battle can offer—which still involves Jerome going to prison. This blatant injustice reflects just how unfair the legal system was for Black people in the 1960s—and how little it did to defend their rights, even when the law is on their side.\\
    When Jerome is killed, the injustice of the legal system is further illuminated. As Jerome lies dying, several policemen do not stop the shooter with force, and instead try to talk to him. Even after the shooter raises his gun to shoot Pearl, Jeff, a civilian, shoots him. As he does so, the policeman angrily asks: “Why in the hell did you shoot him?”. The police’s rage and inaction reflects how little the legal system does to enforce even the laws in place.\\
    This scene establishes the importance of Overcoming Personal Bias. Jerome’s death makes it clear that even new laws and courtroom triumphs are not enough for true change when people act on their own prejudice. Individual people need to change if racism is going to be overcome.\\
    The novel again presents youth as a solution to combatting personal bias. As Jack gives his speech at the conclusion of the trial, he notes how a woman in the crowd “was looking angrily” at him, but that “her boy’s expression was more muted; he actually appeared to be listening”. In this way, the novel shows that the youth are amenable to change. Although the grown woman is angry at the outcome of the trial, there is hope for the future—her son seems to be internalizing what Jack is saying, forming his own opinions instead of perpetuating his mother’s bias.\\
    Jack and DuBose overcome their own personal bias and agree to start a romantic relationship. Their hesitancy has been two-fold. First, they both reflect throughout the novel on how complicated it would be to be with someone of a different race, with DuBose scolding herself for considering it. Second, their hesitancy comes from the fact that they come from such different backgrounds. However, Jack acknowledges that, just like with the trial, there are things that are worth the fight, and he believes that their relationship is one. He tells DuBose that he is “far tougher than [he] thought [he] would be”, reflecting his newfound internal strength and transformation.\\
    
    \bottomrule
    
    \end{tabularx}
\caption{Data examples of book analysis written by literary experts, sourced from the 2024 novel \textit{"A Calamity of Souls."}}
\label{tab:case_3}
\end{table*}

\begin{table*}[t]
  \centering
  \small
    \begin{tabular}{l}
    \toprule
    \rowcolor[gray]{0.95}\multicolumn{1}{c}{\textbf{Prompt I}} \\
    \makecell[l]{
Your task is to help me identify a significant decision point for a character in a book. I will provide you with some\\ information, and you need to return outputs as required.\\\\
\color[rgb]{0,0.39,0}{\# Requirements:}\\
1. The decision must be a life choice of the character, which can reflect the character's personality, past experiences, and\\ interpersonal relationships.\\
2. The decision must have a rationale that can be found earlier in the text, which might be determined by the character's\\ overall personality or by a subtle hint.\\
3. The decision is determined by earlier text, not revealed by reasons in later sections.\\\\
Below are the inputs I will provide and the outputs you need to return.\\
\color[rgb]{0,0.39,0}{\# Inputs:}\\
1. Input 1: A character description written by a human literature expert\\ <description>\\
2. Input 2: The book divided into chapter summaries\\ <chapter>\\
3. Input 3: A book analysis written by a human literature expert\\ <analysis>\\\\
Below is the content you need to output.\\
\color[rgb]{0,0.39,0}{\# Outputs:}\\
1. Output 1: The location of the character's decision point. Please answer with the original text from the chapter\\ summaries (Input 2).\\
Output format: \{"summary\_location":<content>\}\\
2. Output 2: The motivation for the character's decision.\\
Output format: \{"motivation":<content>\}\\
3. Output 3: Chapter numbers related to this decision.\\
Output format: \{"related\_chapter":["chapter\_1",...]\}\\\\
\color[rgb]{0,0.39,0}{\# Execution Steps:}\\
1. Read all inputs.\\
2. Consider the life choice that best reflects the character's personality, past experiences, and interpersonal relationships.\\
3. Output the location of the character's decision point, sourced from the chapter summaries (Input 2).\\
4. Output the motivation for the character's decision.\\
5. Based on the motivation for the character's decision, find the relevant chapters and output the chapter numbers related to\\ this decision.}\\
    \bottomrule
    \end{tabular}
  \caption{Prompt templates for selecting decision points.}
  \label{tab:template_for_select_decsion_point}
\end{table*}
\begin{table*}[t]
  \centering
  \small
    \begin{tabular}{l}
    \toprule
    \rowcolor[gray]{0.95}\multicolumn{1}{c}{\textbf{Prompt II}} \\
    \makecell[l]{
Your task is to locate the position of a given segment of text within the original text. The text segment I provide to you\\ comes from a summary of the original text. I will provide you with the summary and the original text, and you need to\\ find where this segment occurs in the original text. This position must be the exact wording from the original text.\\\\
\color[rgb]{0,0.39,0}{\# Inputs:}\\
1. Input 1: Summary of the original text\\ <summary>\\
2. Input 2: Text from the summary of the original text\\ <text>\\
3. Input 3: Original text\\ <original>\\\\
\color[rgb]{0,0.39,0}{\# Outputs:}\\
Text from the summary of the original text:}\\
    \bottomrule
    \end{tabular}
  \caption{Prompt templates for locate the position of the node in the original book.}.
  \label{tab:template_for_location}
\end{table*}

\begin{table*}[t]
  \centering
  \small
    \begin{tabular}{l}
    \toprule
    \rowcolor[gray]{0.95}\multicolumn{1}{c}{\textbf{Prompt III}} \\
    \makecell[l]{
Your task is to create a multiple-choice question where the correct answer is a decision made by a character in a book.\\ You need to design three incorrect answers. Below are the detailed requirements:\\\\
\color[rgb]{0,0.39,0}{\# Requirements:}\\
1. Provide the scenario in which the character is making the decision.\\
2. Design three incorrect answers that are reasonable and could be choices the character might make, but are not the\\ optimal choice.\\
3. Ensure that there is no data leakage in any of the outputs.\\\\
\color[rgb]{0,0.39,0}{\# Inputs:}\\
1. Input 1: Character description written by a human literature expert\\ <description>\\
2. Input 2: Summary of the entire book divided by chapters\\ <chapter>\\
3. Input 3: Book analysis written by a human literature expert\\ <analysis>\\
4. Input 4: Location of the character's decision point\\ <location>\\
5. Input 5: Motivation for the character's decision\\ <motivation>\\
6. Input 6: Original text from chapters related to the decision\\ <original>\\\\
\color[rgb]{0,0.39,0}{\# Outputs:}\\
1. Output 1: Scenario in which the character is situated.\\
Output format: \{"scenario":<content>\}\\
2. Output 2: Multiple-choice question.\\
Output format: \{"question":<q>,"options":\{"A":<o1>,"B":<o2>,"C":<o3>,"D":<o4>\}\}\\\\
\color[rgb]{0,0.39,0}{\# Execution Steps:}\\
1. Read all inputs.\\
2. Output the scenario in which the character faces this decision.\\
3. Output the multiple-choice question, ensuring that the incorrect options are also reasonable.}\\
    \bottomrule
    \end{tabular}
  \caption{Prompt templates for constructing multiple-choice questions.}
  \label{tab:template_for_qa_construction}
\end{table*}
\begin{table*}[t]
  \centering
  \small
    \begin{tabular}{l}
    \toprule
    \rowcolor[gray]{0.95}\multicolumn{1}{c}{\textbf{Prompt IV}} \\
    \makecell[l]{
Please play the role of <Character A> based on the <Profile> and make your life choice under the <Scenario> regarding\\ <Question>. Return the option letter (A, B, C, or D) that your character should most appropriately choose in the current\\ scenario. The <Profile> consists of <Description> and <Memory>, where <Description> is an overall description of the\\ character, and <Memory> consists of specific events the character has experienced.\\\\
\color[rgb]{0,0.39,0}{\# Inputs:}\\
1. Profile:\\ 
1.1. Description\\
<description>\\
1.2. Memory\\
<memory>\\
2. Scenario:\\ 
<scenario>\\
3. Question:\\ 
<question>\\
4. Options:\\
<option>\\\\
\color[rgb]{0,0.39,0}{\# Outputs:}\\
Your choice(A, B, C, or D):}\\
    \bottomrule
    \end{tabular}
  \caption{Prompt templates for role-playing as the character.}
  \label{tab:template_for_system_prompt}
\end{table*}

\begin{table*}[t]
  \centering
  \small
    \begin{tabular}{l}
    \toprule
    \rowcolor[gray]{0.95}\multicolumn{1}{c}{\textbf{Prompt V}} \\
    \makecell[l]{
Your task is to find segments within a character's <Description> that may relate to the content of the <Scenario> and <Question>.\\ The <Scenario> describes the situation the character is in, and the <Question> asks what choice the character should make. The\\ segments you need to find could influence the character's motivation for making their choice, including aspects that shape the\\ character's personality and foreshadowing related to the decision scenario.\\\\
\color[rgb]{0,0.39,0}{\# Inputs:}\\
1. Description:\\ 
<description>\\
2. Scenario:\\ 
<scenario>\\
3. Question:\\ 
<question>\\\\
\color[rgb]{0,0.39,0}{\# Outputs:}\\
Segments that may influence the character's choice:}\\
    \bottomrule
    \end{tabular}
  \caption{Prompt templates for \method.}
  \label{tab:template_for_method}
\end{table*}

\begin{table*}[t]
  \centering
  \small
    \begin{tabular}{l}
    \toprule
    \rowcolor[gray]{0.95}\multicolumn{1}{c}{\textbf{Manual Examination Rules}} \\
    \makecell[l]{
\color[rgb]{0,0.39,0}{\textbf{1. Comprehensiveness}}\\
\textbf{Rule:}\\
Evaluators must ensure that each multiple-choice question fully considers the character's background, context, and motivation.\\ The questions should reflect the true decisions and experiences of the character within the narrative.\\
\textbf{Scoring Guide:}\\
\textit{Score 2 (Excellent): The question is detailed and comprehensive, aligning perfectly with the character's background and}\\ \textit{motivation.}\\  
\textit{Score 1 (Average): The question aligns generally but is missing key aspects of the character's background information or}\\ \textit{motivational nuances.}\\
\textit{Score 1 (Poor): The question significantly misaligns with the character's background or motivation.}\\
\\
\color[rgb]{0,0.39,0}{\textbf{2. Logical Consistency}}\\
\textbf{Rule:}\\
Evaluators should assess the internal consistency and plausibility of the question within the narrative thread. The content and\\ structure of the multiple-choice question must be consistent with the plot and the character's logical decision-making process.\\
\textbf{Scoring Guide:}\\
\textit{Score 2 (Excellent): The question is entirely consistent with the character's known decisions and the structure of the plot.}\\
\textit{Score 1 (Average): The question is generally consistent but has minor inconsistencies in detail.}\\
\textit{Score 0 (Poor): The question is logically inconsistent with the character's known decisions or the structure of the plot.}\\
\\
\color[rgb]{0,0.39,0}{\textbf{3. Challenge Level}}\\
\textbf{Rule:}\\
Evaluators need to assess the plausibility of the incorrect options. Wrong options should be reasonably believable and\\ attractive within the constraints of the character's background and motivations, making the questions sufficiently challenging.\\
\textbf{Scoring Guide:}\\
\textit{Score 2 (Excellent): All incorrect options are highly plausible and convincingly misleading.}\\
\textit{Score 1 (Average): Most incorrect options are reasonable, but one or two lack plausibility.}\\
\textit{Score 0 (Poor): Incorrect options are obviously illogical and lack the ability to mislead.}\\
\\
\color[rgb]{0,0.39,0}{\textbf{4. Alignment with Character Motivation}}\\
\textbf{Rule:}\\
Evaluators must assess whether the question correctly guides the testing model to step into the role and make a choice, i.e.,\\ testing if the model can replicate the real storyline's choices. It is crucial that the character's motivations, as articulated by\\ literary experts, are a central component reflected in these questions.\\
\textbf{Scoring Guide:}\\
\textit{Score 2 (Excellent): The question unambiguously points to a specific character decision point, accurately testing the model's}\\ \textit{ability to role-play.}\\
\textit{Score 1 (Average): The question points to a character decision point to some extent, but the indicators are not clear enough,}\\ \textit{potentially reducing the accuracy of the model's role-playing test.}\\
\textit{Score 0 (Poor): The question fails to clearly define the character decision point, unable to test the model's role-playing ability}\\ \textit{effectively.}\\
\\
\textbf{Additional Notes:}\\
1. Before starting the evaluation, each evaluator must understand the core motives and development axes of the character by\\ reading summaries and analyses of the novels created by literary experts.\\
2. Ensure that evaluators are familiar with all background material before scoring any questions.\\
3. Evaluators should reference the analyses by literary experts of the characters to evaluate each of GPT-4's multiple-choice\\ questions, maintaining consistency of standards.\\
4. Application of the evaluation rules should be flexible and adapted to the specific context; scoring standards may be adjusted\\ for special cases.\\}\\
    \bottomrule
    \end{tabular}
  \caption{Guidelines for Manual Examination of Multiple-Choice Questions in Literary Analysis.}
  \label{tab:manual_examination_rules}
\end{table*}

\section{Prompts}
\label{sec:appendix_prompt}
In this section, we provide the key prompts we used, including prompts for selecting decision points, locating the node's position, constructing multiple-choice questions, system prompts for role-playing characters, and the prompt of \method.

\subsection{Selecting Decision Points}
\label{sec:select}
As mentioned in section \ref{sec:dataset_construction}, the first step in constructing our data is selecting the character's Decision Points. In this step, our input data consists of all raw data from \textit{Supersummary}, including the current character's description, all chapter summaries, and book analysis. We provide this data to GPT-4, requesting it to output three types of data: the character's decision point, the corresponding gold motivation, and the chapters related to this choice. 
Table \ref{tab:case_1} presents the character's description, Table \ref{tab:case_2} presents the chapter summaries, and Table \ref{tab:case_3} presents the book analysis, these examples are all from the 2024 novel \textit{"A Calamity of Souls."}. Table \ref{tab:template_for_select_decsion_point} illustrates the prompt for selecting decision points.

\subsection{Locating the Node}
Furthermore, for the character's decision points, we provide the previously identified decision points and their corresponding chapters from the original book to GPT-4. This allows it to precisely determine the position in the original book that should be segmented, helping to avoid data leakage. Table \ref{tab:template_for_location} shows the prompt for locating.

\subsection{Constructing Multiple-Choice Questions}
After selecting the Decision Points, our next step is constructing Multiple-Choice Questions. In this step, our input data consists of all the input and output data from the previous step(Appendix \ref{sec:select}). We ask GPT-4 to construct a multiple-choice question regarding the character's decision based on this data, outputting the scenario in which the character is situated, the question, and four options. Table \ref{tab:template_for_qa_construction} shows the prompt for constructing multiple-choice questions.

\subsection{System Prompts for Role-Playing}
The prompt for role-playing as the character can be found in Table \ref{tab:template_for_system_prompt}.

\subsection{Prompts for \method}
The prompt for \method can be found in Table \ref{tab:template_for_method}.

\section{Dateset Details}
\subsection{Categories of novel}
\label{appendix:category_1}
Below is a complete classification of novel genres, from the literary experts at the Supersummary website:

\textbf{Mystery Novels}: The mystery genre includes general mystery, noir mystery, historical mystery, police procedural mystery, and supernatural mystery.

\textbf{Thriller Novels}: The thriller genre includes supernatural thrillers, historical thrillers, environmental thrillers, medical thrillers, legal thrillers, political thrillers, military thrillers, and espionage stories.

\textbf{Science Fiction Novels}: Science fiction stories take place in the future or the past but are almost always set in a dimension different from our present. They are characterized by entirely new, imagined realities and universes, where the setting is indispensable. High technology also plays an important role in these stories. Space opera, romantic science fiction, military science fiction, alternate history, dystopian and utopian tales, as well as steampunk, are considered sub-genres of science fiction.

\textbf{Romance Novels}: Romance novels feature romantic relationships between at least two people, characterized by tension and desire. Romance novel themes include supernatural romance, contemporary romance, historical romance, western romance, gothic romance, regency romance, and romantic suspense.

\textbf{Fantasy Novels}: Fantasy stories are centered around mythical kingdoms and magic. Fantasy novel genres include contemporary fantasy, traditional fantasy, horror fantasy, weird fantasy, epic fantasy, historical fantasy, dark fantasy, urban fantasy, and anime fantasy.

\textbf{Action Adventure Novels}: Action-adventure novels place the protagonist in various realistic dangers. This is a fast-paced genre where the climax should provide some form of thrill for the audience or reader.

\textbf{Speculative Novels}: Speculative fiction is characterized by overlapping with our world but differing in key aspects, introducing "what if" scenarios.

\textbf{Mystery Thriller Novels}: Mystery thriller stories are usually filled with suspense, with one or more characters' lives in danger. In gripping scenes, these characters are often chased and manage to escape narrowly.

\textbf{Young Adult Novels}: Young Adult fiction, commonly abbreviated as YA, is intended for teenagers aged 12-18. Most YA novels feature coming-of-age stories, often with elements of science fiction or fantasy.

\textbf{New Adult Novels}: New Adult novels target college-aged adults and usually explore stories of first adventures on one's own.

\textbf{Horror and Supernatural Novels}: Horror, supernatural, and ghost story genres aim to scare the reader and audience by playing on common fears. The protagonist usually has to overcome supernatural threats, and the stories often include supernatural elements.

\textbf{Crime Mystery Novels}: Crime mystery stories focus on a central problem or crime to be solved, or a mysterious event that must be answered. Throughout the story, the reader or audience and characters are given clues that help the protagonist eventually find the solution.

\textbf{Detective Novels}: In detective fiction, a common element is a police officer or detective embarking on solving a crime. The plot is filled with evidence gathering, forensic studies, and legal drama.

\textbf{Historical Novels}: Historical novels are fictional stories set against the backdrop of real historical events or historical settings. Historical fiction may also portray real historical figures.

\textbf{Western Novels}: Stories with a western theme take place in the old times of the American West, filled with adventure, cowboys, and pioneers. There are also Italian western novels, Asian western novels, space westerns, and other stories about the American West.

\textbf{Family Saga Novels}: Family saga novels typically tell the stories of several generations of family members dealing with family affairs, family curses, and family adventures. These stories usually follow a timeline and deal with conflicts in the present.

\textbf{Women's Novels}: Women's fiction plotlines revolve around the challenges and crises that women face in real life, including interpersonal relationships, work, family, politics, and religion.

\textbf{Magical Realism Novels}: Magical realism stories take place in the real world but have characters who take magical elements for granted. These mystical elements do not exist in real life, but they are perfectly normal in magical realism.

\subsection{Categories of motivations}
\label{appendix:category_2}
Below are the motivations for each topic and their corresponding proportions:

\paragraph{Character-driven motivation}
Character-driven narrative is centered on the inner world, growth, and transformation of characters. In character-driven stories, the progression of the plot and the resolution of conflicts are often propelled by the characters' personalities, desires, fears, and psychological development. Such stories typically delve deeply into the characters' mental states and development, focusing on how characters influence each other and how their actions reflect their inner emotions and thoughts. The choices and changes of the characters serve as the main engine for the story's development, influencing the direction of the plot. Sub-motivations of character-driven behavior include:

\textbf{Personality and Traits}: (27.12\%) These refer to a character's characteristics such as being introverted, extroverted, brave, or guilt-ridden, which influence their choices and lifestyle.

\textbf{Emotions and Psychological State}: (7.53\%) A character's emotional responses, psychological traumas, or sense of personal well-being are key elements that drive the story forward.

\textbf{Social Relationships}: (6.31\%) The character's status and changes in family, love, friendship, or other social connections can propel the story's development.

\textbf{Values and Beliefs}: (27.12\%) The character's moral convictions, religious beliefs, or life philosophy can serve as motivation for action.

\textbf{Desires and Goals}: (7.22\%) Personal desires, career aspirations, or specific life goals of a character are pivotal in advancing the plot.

\paragraph{Plot-driven motivation}
Plot-driven narrative emphasizes the creation and resolution of external conflicts in the story. In such stories, the driving force of the plot comes from a series of events and conflicts themselves, while characters are often the responders to these events. Plot-driven stories typically highlight tense drama, complex plot structure, and frequent changes in external actions, rather than changes in the character's internal world. In this type of narrative, characters may act in response to the demands of the plot, rather than the plot following the development of the characters' inner world. Sub-motivations of plot-driven behavior include:

\textbf{External Conflicts}: (8.76\%) Conflicts from the outside world, such as war, natural disasters, or social upheaval, can propel the plot.

\textbf{Tasks and Goals}: (4.7\%) Tasks or specific goals that characters must accomplish often become the driving force behind the story's progression.

\textbf{Puzzles and Secrets}: (7.22\%) Secrets that need revealing or mysteries that need solving can form the core of a story.

\textbf{Pursuits and Escapes}: (4.25\%) Characters might chase something (e.g., power, wealth, knowledge) while avoiding or fleeing from certain situations (e.g., pursuit, personal past).

\textbf{Exploration and Discovery}: (3.66\%) Characters' adventures or discoveries in new realms (physical, scientific, or spiritual) can move the plot forward.

\textbf{Power and Control}: (4.81\%) The pursuit or struggle for power and control often serves as motivation for characters.

\textbf{Intrigue and Betrayal}: (4.09\%) Complex plots and betrayals can catalyze the progression of the story.

\section{Manual Annotation}
\label{sec:annotation}
This is a supplement to Section \ref{sec:dataset_construction}. After constructing the multiple-choice question data using GPT-4, we perform a manual examination. For each annotator, we provide key character descriptions, chapter summaries, and book analyses written by human literature experts on the \textit{Supersummary}. Each annotator is asked to score the questions constructed by GPT-4 based on the rules shown in Table \ref{tab:manual_examination_rules}. We evaluated the scores of each annotator and only retained the data with an average score of more than 6 points.

We provide compensation based on the local minimum hourly wage for all individuals involved in the annotation.

\section{Future direction}
Building personal agents for everyone is an exciting topic. We have explored how fictional characters can determine their subsequent actions based on historical data, proposing possibilities for combining role-playing with personalized models. We believe there are additional directions to explore:
\begin{itemize}
\item \textbf{Real-life version of \dataset} The behavior of fictional characters often stems from the author's design, which can lead to logical inconsistencies. In contrast, real-world human behavior data do not have this issue. How to construct real human historical data and related behavior data is a question worth exploring.

\item \textbf{Improving RPLA performance in life choices} Although \method has achieved decent results, better methods are needed to balance reasoning efficiency (which depends on the length of input tokens) while achieving superior outcomes.

\item \textbf{More complex downstream decision tasks} The decisions we select are often significant choices for fictional characters, resulting in a large decision space without a fixed task framework. Identifying more systematic tasks by integrating social sciences is a challenge that needs to be addressed in the future.
\end{itemize}

\end{CJK}
\end{document}